\documentclass{sig-alternate}
\usepackage{mathptmx} 
\usepackage{paralist}
\usepackage{fancyhdr}
\usepackage{multirow}
\usepackage{subfigure}
\usepackage{algorithm}
\usepackage{algorithmic}
\usepackage{enumitem}
\usepackage[normalem]{ulem}
\usepackage[hyphens]{url}
\usepackage[sort,nocompress]{cite}
\usepackage[final]{microtype}
\usepackage[keeplastbox]{flushend}
\usepackage[bookmarks=true,breaklinks=true,letterpaper=true,colorlinks,linkcolor=black,citecolor=blue,urlcolor=black]{hyperref}

\fancypagestyle{firstpage}{
  \fancyhf{}
  
  \fancyhead[C]{\vspace{15pt}\normalsize{Under Review}} 
  \fancyfoot[C]{\thepage}
}

\title{AutoScale: Optimizing Energy Efficiency of End-to-End Edge Inference under Stochastic Variance 
\vspace{-1.2cm}
}

\author{\normalsize Young Geun Kim$^{\ast}$ and Carole-Jean Wu$^{\ast \dagger}$\\ 
\normalsize Arizona State University$^{\ast}$    Facebook AI$^{\dagger}$\\
 \normalsize younggeun.kim@asu.edu \hfill{      } carolejeanwu@fb.com
}

\begin{document}
\maketitle
\thispagestyle{firstpage}
\pagestyle{plain}


\begin{abstract}
Deep learning inference is increasingly run at the edge. As the programming and system stack support becomes mature, it enables acceleration opportunities within a mobile system, where the system performance envelope is {\it scaled up} with a plethora of programmable co-processors. Thus, intelligent services designed for mobile users can choose between running inference on the CPU or any of the co-processors on the mobile system, or exploiting connected systems, such as the cloud or a nearby, locally connected system. By doing so, the services can {\it scale out} the performance and increase the energy efficiency of edge mobile systems. This gives rise to a new challenge--deciding {\it when} inference should run {\it where}. Such {\it execution scaling decision} becomes more complicated with the stochastic nature of mobile-cloud execution, where signal strength variations of the wireless networks and resource interference can significantly affect real-time inference performance and system energy efficiency. To enable accurate, energy-efficient deep learning inference at the edge, this paper proposes \textit{AutoScale}. \textit{AutoScale} is an adaptive and light-weight execution scaling engine built upon the custom-designed reinforcement learning algorithm. It continuously learns and selects the most energy-efficient inference execution target by taking into account characteristics of neural networks and available systems in the collaborative cloud-edge execution environment while adapting to the stochastic runtime variance. Real system implementation and evaluation, considering realistic execution scenarios, demonstrate an average of 9.8 and 1.6 times energy efficiency improvement for DNN edge inference over the baseline mobile CPU and cloud offloading, while meeting the real-time performance and accuracy requirement. 
\end{abstract}
\vspace{-0.2cm}
\section{Introduction}
\label{sec:introduction}

It is expected that there will be more than 7 billion mobile device users and 900 million wearable device users in 2021~\cite{Statista,Statista2}, including smartphones, smart watch, wearable virtual or mixed reality devices. To improve mobile user experience, various intelligent services, such as virtual assistance~\cite{Amazon,Apple}, face/image recognition~\cite{Google_1}, and language translation~\cite{Google_2}, have been introduced in recent years. Many companies, including Amazon, Facebook, Google, and Microsoft, are using sophisticated machine learning models, especially Deep Neural Networks (DNNs) as the key machine learning component for these intelligent services~\cite{Amazon,Google_2,Microsoft,CJWu2019}.

Traditionally, due to the compute- and memory-intensive nature of the DNN workloads \cite{CDing2017,JHauswald2015,SBianco2018}, both training and inference were executed on the cloud~\cite{YKang2017,AEEshratifar2018}, while the mobile devices only acted as user-end sensors and/or user interfaces. More recently, with the advancements of powerful mobile System-on-Chips (SoCs)~\cite{MHalpern2016,LNHuynh2017,SWang2020_2}, there have been increasing pushes to execute DNN inference on the edge mobile devices~\cite{ECai2017,AEEshratifar2018,MHan2019,YKang2017,YKim2019,NDLane2016,SWang2020_1,SWang2020_2,CJWu2019,GZhong2019}. This is because executing inference at the edge can improve the response time of services, by removing data transmission overhead. However, executing inference on the edge mobile devices also results in increased energy consumption of the mobile SoCs~\cite{YKang2017}. Since the edge mobile devices are energy-constrained~\cite{YGKim2018}, it is necessary to optimize the energy efficiency of the DNN inference, while satisfying the Quality-of-Service (QoS) requirements of these services.

To address these performance and energy efficiency challenges, modern mobile devices employ more and more accelerators and/or co-processors, such as Graphic Processing Units (GPU), Digital Signal Processors (DSPs), and Neural Processing Units (NPUs)~\cite{TChen2018,AIgnatov2018}, \textit{scaling up} the overall system performance. Furthermore, the mobile system stack support for DNNs has become more mature, allowing DNN inference to leverage the computation and energy efficiency advantages provided by the co-processors. For example, modern deep learning compiler and programming stacks, such as TVM~\cite{TChen2018}, SNPE~\cite{Qualcomm}, and Android NN API~\cite{Android,AIgnatov2018}, enable inference execution on a diverse set of hardware back-ends.

These recent advancements give rise to a \textit{new} challenge---deciding \textit{when} inference should run \textit{where}. 
Intelligent services aiming to run on the mobile devices can choose between running inference on the CPU or any of the co-processors on the device, or exploiting connected systems, such as the cloud or a nearby, locally-connected system \cite{JIBenedetto2019} that is more powerful than the device itself. By doing so, the services can {\it scale out} the performance and increase the energy efficiency of edge mobile devices. For example, many personalized health and entertainment use cases are powered by a collaborative execution environment composed of smart watches, smartphones, and the cloud~\cite{CH,Fitbit,OMRON,WITHINGS}. Similarly, virtual and augmented reality systems consist of wearable electronics, smartphones as the staging device, and the cloud~\cite{Google_3, Google_4, Microsoft2, Oculus}. However, the decision process is challenging for any intelligent services, since energy efficiency of each execution target significantly vary depending on various features, such as NN characteristics and/or edge-cloud system profiles. The extremely fragmented mobile SoCs make this decision process even more difficult, as there are myriads of hardware targets with different profiles \cite{CJWu2019} to choose from.

To determine the optimal execution scaling decision, state-of-the-art approaches, such as~\cite{AEEshratifar2018,MHan2019, YKang2017,SWang2020_1,SWang2020_2,GZhong2019}, proposed to build predictive models. However, these prior approaches did not consider stochastic runtime variances, such as interference from co-running tasks or network signal strength variations, which have a large impact on energy efficiency~\cite{BGaudette2019}. In a realistic execution environment, there can be several applications simultaneously running along with the DNN inference~\cite{YGKim2017_1,SYLee2017,DShingari2018}, since recent mobile devices support multitasking features~\cite{DShingari2018} such as screen sharing between multiple applications. In addition, signal strength variations of the wireless networks can significantly affect performance and energy efficiency of cloud inference, since the data transmission latency and energy exponentially increase when the signal strength is weak~\cite{YGKim2019}, which accounts for 43\% of data transmission~\cite{NDing2013}. Therefore, without considering such stochastic variances, one would not be able to choose the optimal execution scaling decision for DNN inference.

This paper proposes an \textit{adaptive} and \textit{light-weight} execution scaling engine, called {\em AutoScale}, to make accurate scaling decisions for the {\em optimal execution target} of edge DNN inference {\em under the presence of stochastic variances}. Since the optimal execution target significantly varies depending on the NN characteristics, the underlying execution platforms, as well as the stochastic runtime variances, it is infeasible to enumerate the massive design space exhaustively. Therefore, \textit{AutoScale} leverages a lightweight reinforcement learning technique for continuous learning, that captures and adapts to the environmental variances of stochastic nature ~\cite{BDonyanavard2019,VMnih2015,RNishtala2017,HShen2013}. {\em AutoScale} observes NN characteristics, such as layer composition, and current system information, such as interference intensity and network stability, and selects an execution target which is expected to maximize the energy efficiency of DNN inference, satisfying the performance and accuracy targets. The result of the selection is then measured from the system and fed back to \textit{AutoScale}, allowing \textit{AutoScale} to continuously learn and predict the optimal execution target. We demonstrate {\em AutoScale} with real system-based results that show improved energy efficiency of DNN inference by 9.8X and 1.6X on average, compared to the baseline settings of mobile CPU and cloud offloading, satisfying both the QoS and accuracy constraints with 97.9\% of prediction accuracy.

This paper makes the following key contributions:
\begin{compactitem}
    \item This paper provides an in-depth characterization of DNN inference execution on mobile and edge-cloud systems. The characterization results show that the optimal execution scaling decision significantly varies depending on the NN characteristics and the stochastic nature of mobile execution (Section \ref{sec:motivation}).
    \item This paper proposes an intelligent execution scaling engine that accurately selects the optimal execution target of mobile inference in the presence of stochastic variances (Section \ref{sec:design}).
    \item To demonstrate the feasibility and practicality of the proposed execution scaling engine, we implement and evaluate \textit{AutoScale} with a variety of on-device inference use cases under the edge-cloud execution environment using real systems and devices, allowing \textit{AutoScale} to be adopted immediately\footnote{We plan to open source \textit{AutoScale} upon paper acceptance.}  (Section \ref{sec:result}).
\end{compactitem}

\vspace{-0.2cm}
\section{Background}
\label{sec:background}
This section introduces the necessary background that makes up the components for the \textit{AutoScale} framework, i.e., DNN, inference at the edge, and QoE of real-time inference.
\vspace{-0.1cm}
\subsection{Deep Neural Network}
\label{sec:background1}

DNNs are constructed by connecting a large number of functional layers to extract features from inputs at multiple levels of abstraction~\cite{AKarpathy2014,HLee2009}. Each layer is composed of multiple processing elements (\textit{neurons}), which are applied with the same function to process different parts of an input. Depending on what function is applied, the layers can be classified into the various types~\cite{CDing2017}. These layers and their execution characteristic differences are essential since they can affect the decision made by \textit{AutoScale}. We give brief descriptions for each layer type below.

{\it Convolutional layer} ({\bf CONV}) performs a two-dimensional convolution to extract a set of feature maps from its input. To selectively activate meaningful features, an activation function, such as sigmoid or rectified-linear, is applied to the obtained feature maps. Typically, this layer is compute-intensive due to the calculation of convolutions. 

{\it Fully-connected layer} ({\bf FC}) computes the weighted sum of the inputs using a set of weights and then applies the activation function to the weighted sum of the inputs. This layer is one of the most compute- and memory-intensive layers in DNNs \cite{CDing2017,YKang2017,YKim2019}, since its neurons are exhaustively connected to all the neurons in the previous layer.

{\it Recurrent layer} ({\bf RC}) is a layer where the output of current step in a sequence is used as an additional input in the next step of the sequence. In each step, this layer also computes the weighted sum of the inputs using a set of weights. This layer is even more compute- and memory-intensive than FC layer, since its neurons can be connected to neurons in the previous, current, and the next layer.

Other commonly-used layers include:
{\it Pooling layer} applies a sub-sampling function, such as max or average, to regions of the input feature maps;
{\it Normalization layer} normalizes features across spatially grouped feature maps;
{\it Softmax layer} produces a probability distribution over the number of possible classes for classification; 
{\it Argmax layer} chooses the class with the highest probability;
{\it Dropout layer} randomly ignores neurons during training and allows the neurons to pass through during inference.
These layers are typically less compute- and memory-intensive than CONV, FC, and RC layers, such that they do not have a large impact on performance and energy efficiency of DNN inference. 

DNNs can be constructed with various compositions of layers. For example, NNs used for computer vision applications (e.g., Inception, Mobilenet, Resnet, etc.) are mainly composed of CONV, POOL and FC layers. On the other hand, NNs used for language processing applications (e.g., BERT) mainly consist of RC layers, such as Long Short-Term Memory (LSTM) and attention. Since each layer has unique characteristics due to different compute and memory intensities, to optimize inference execution for DNNs, it is important to consider the layer composition.
\vspace{-0.1cm}
\subsection{DNN Inference Execution at the Edge}
\label{sec:background2}

\begin{figure}[t]
    \centering
    \includegraphics[width=\linewidth]{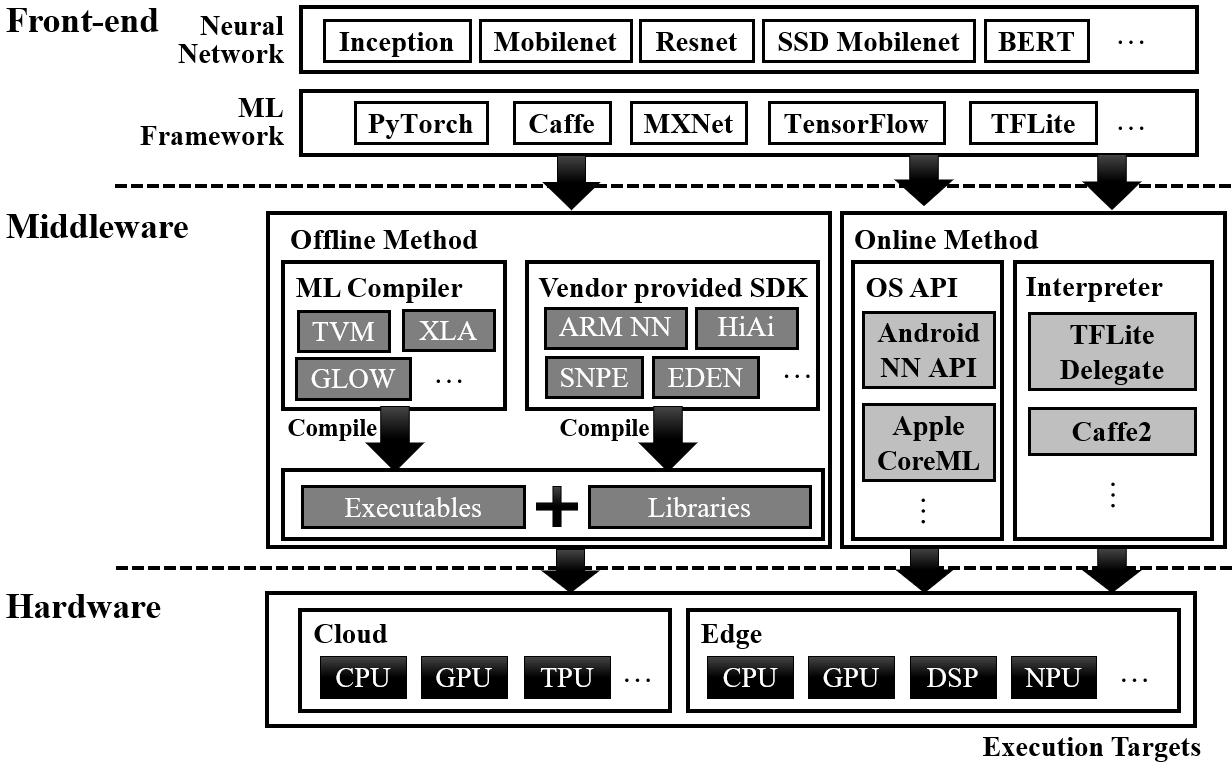}
    \vspace{-0.7cm}
    \caption{System stack for DNN inference execution.}
    \label{fig:system stack}
    \vspace{-0.5cm}
\end{figure}

Figure~\ref{fig:system stack} depicts the general structure of the system stack for machine learning inference execution at the edge. At the front-end, DNNs are implemented with various frameworks, such as TensorFlow~\cite{Tensorflow}, PyTorch~\cite{PyTorch}, Caffe~\cite{Caffe}, MXNet~\cite{MXNet}, etc, whereas the middleware allows deployment of DNN inference execution onto a diverse set of hardware back-ends. They also enable efficient inference at the edge---various NN optimizations, such as quantization \cite{MCourbariaux2016,JFromm2020,BJacob2018,JHKo2017,NDLane2016,CJWu2019,RZhao2019}, weight compression~\cite{SHan2016,DLi2018} and graph pruning \cite{JYu2017,TZhang2018} can be employed before the DNNs are deployed. Among the optimizations, the quantization is one of the most widely used ones for the edge execution, since it reduces both compute and memory intensities of the inference; quantization shrinks the 32-bit floating-point values (FP32) of NNs to fewer bits such as 16-bit FP values (FP16) or 8-bit integer values (INT8). Since the middleware does not select a specific hardware target for DNN inference execution, intelligent services should choose one among the possible hardware targets. However, this decision process is challenging, since energy efficiency of each execution target can significantly vary depending on various features. 

\vspace{-0.1cm}
\subsection{Real-Time Inference Quality of Experience}
\label{sec:background3}

Quality of user experience is a key metric for mobile optimization. For real-time inference, the Quality-of-Experience (QoE) is the product of inference latency, inference accuracy, and system energy efficiency.
To improve energy efficiency of mobile devices, a number of energy management techniques can be used~\cite{YGKim2018}. Unfortunately, the techniques often sacrifice performance (i.e., latency) for energy efficiency, degrading QoE of real-time inference.

Inference latency is an important factor for QoE, since if the latency of a service exceeds the human acceptable limit, users would abandon the service~\cite{DShingari2018,YZhu2015}. However, a single-minded pursuit of performance is not desirable in mobile devices due to their energy constrained nature. Hence, there is a need to provide just enough performance to meet the QoS expectations of users with minimal energy consumption. The QoS expectation of users can be defined as a certain latency value (e.g., 33.3 ms for 30 FPS video frame rate~\cite{BEgilmez2017,YZhu2015} or 50 ms for interactive applications~\cite{YEndo1996,DLo2015}), below which most users cannot perceive any notable difference.

Various NN optimizations can improve both the latency and energy efficiency of inference. However, the optimizations often sacrifice inference accuracy. Since human-level accuracy is one of the key requirements toward user satisfaction~\cite{SBianco2018,CDing2017,YKim2019}, it is also important to maintain the inference accuracy above the inference quality expectation of users.  

In summary, to maximize the quality of user experience for real-time inference, it is crucial to maximize the system-wide energy efficiency while satisfying the human acceptable latency and accuracy expectations.
\vspace{-0.2cm}
\section{Motivation}
\label{sec:motivation}

This section presents system characterization results for realistic DNN inference scenarios deployed on real mobile and edge-cloud systems. We examine the design space that covers three important axes---latency, accuracy, and energy efficiency (performance per watt). 

For mobile inference, we select three smartphones---Xiaomi Mi8Pro, Samsung Galaxy S10e, and Motorola Moto X Force---to represent the categories of high-end mobile systems with GPU and DSP co-processors, high-end mobile systems with GPU but without DSP, and mid-end mobile systems\footnote{High-end mobile systems with and without an NN-specialized accelerator (i.e., DSP) are used to examine the performance scale-up from off-the-shelf mobile systems. In addition, we select Moto X Force to represent the mid-end mobile systems with a much wider market coverage~\cite{CJWu2019} (see detail in Section \ref{sec:methodology}).}, respectively. The edge-cloud inference execution is emulated with the three smartphones and a server-class Intel Xeon processor, hosting an NVIDIA P100 GPU. For a locally connected mobile device, we use a tablet, Samsung Galaxy Tab S6; note that we connect the smartphones with the tablet via a Wi-Fi-based peer-to-peer wireless network, Wi-Fi direct. Detailed specifications of the mobile and edge-cloud execution setup are presented in Section \ref{sec:methodology}.

\vspace{-0.1cm}
\subsection{Varying Optimal DNN Execution Target}
\label{sec:motivation1}

\begin{figure}[t]
    \centering
    \includegraphics[width=\linewidth]{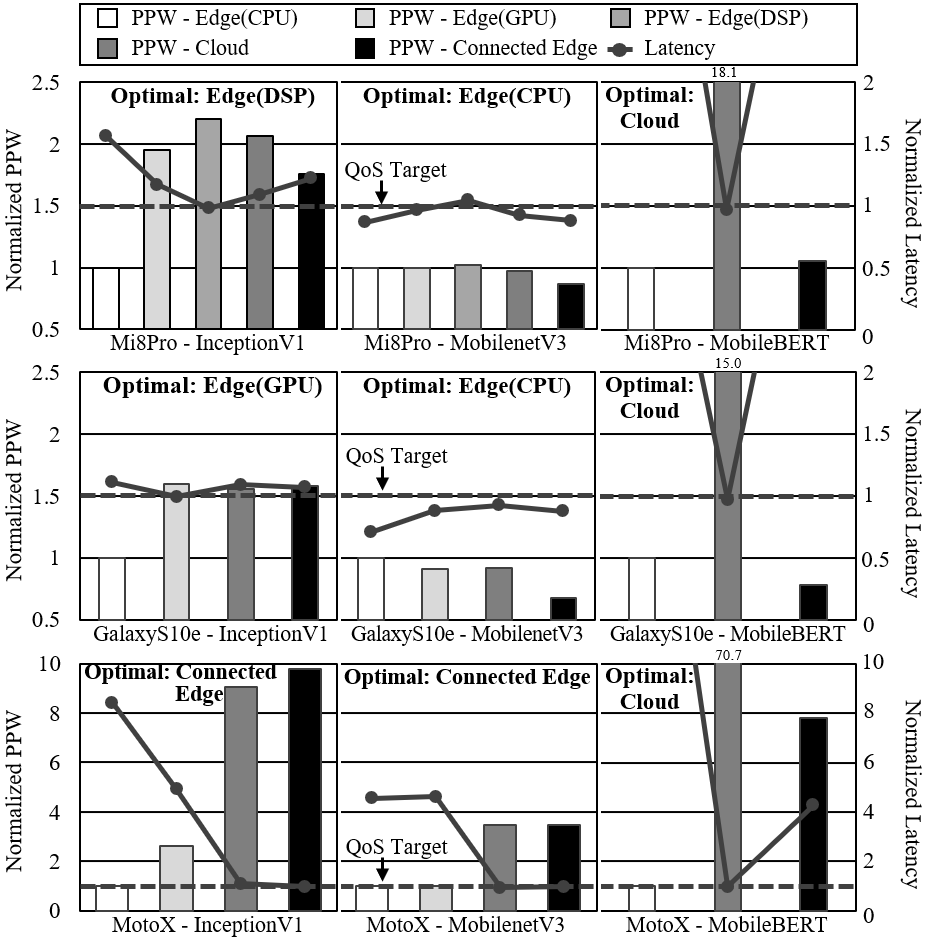}
    \vspace{-0.6cm}
    \caption{Optimal execution target depends on NN characteristics and edge-cloud system profiles. Note that PPW is normalized to Edge(CPU) and latency is normalized to the QoS target.}
    \label{fig:energy vs performance}
    \vspace{-0.4cm}
\end{figure}

\begin{itemize}[leftmargin=*]
\item \textit{Optimal edge-cloud execution depends on the NN characteristics and edge-cloud system profiles.}
\end{itemize}

Figure~\ref{fig:energy vs performance} shows the energy efficiency and latency of three commonly-deployed mobile inference use cases over the three mobile and the edge-cloud setup. The x-axis represents the running mobile system with three representative NNs.

For the high-end systems (i.e., Mi8Pro and Galaxy S10e), the optimal edge-cloud execution shifts, depending on NN characteristics. For example, in the case of light NNs, such as InceptionV1 and MobilenetV3, edge inference execution is more efficient than cloud inference execution. This is because the performance of off-the-shelf mobile SoCs is sufficient to satisfy the QoS target of the light NNs. On the other hand, in the case of heavy NNs, such as MobileBERT, cloud execution is more efficient than edge execution, since the performance of the mobile SoCs is insufficient. In this case, the performance gain of cloud execution (reduced computation time and energy) outweighs its loss (increased data transmission time and energy). 

For the mid-end system (i.e., Moto X Force), however, scaling out to the connected systems is always advantageous, since performance of the SoC in this system is not enough even for the light NNs. In the case of light NNs, scaling out to a locally connected device could be an option, as opposed to scaling out to the cloud, since 1) the higher-end device (i.e., tablet) can satisfy the QoS constraint of the light NNs, and 2) data transmission overhead between the locally connected edge devices is usually smaller than that between edge-cloud. On the other hand, in case of heavy NNs, there is no other option than scaling out to the cloud.

\begin{itemize}[leftmargin=*]
\item \textit{Optimal execution target depends on layer compositions.}
\end{itemize}

\begin{figure}[t]
    \includegraphics[width=\linewidth]{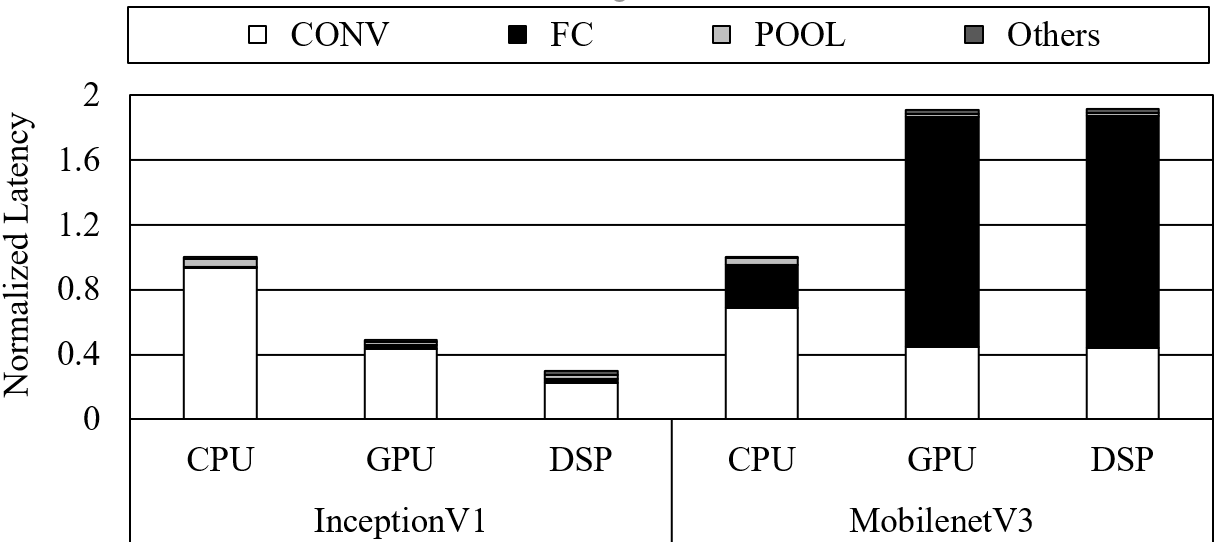} 
    \vspace{-0.6cm}
    \caption{Each layer in NNs exhibits different latency on different mobile processors. For this reason, the optimal execution target for NNs vary depending on layer compositions. Note that latency is normalized to that of CPU.}
    \label{fig:layer compositions}
    \vspace{-0.5cm}
\end{figure}

Another important observation in edge inference execution is that, the optimal execution target can vary depending on the layer compositions of the NNs. Figure~\ref{fig:layer compositions} shows the cumulative latency of different layers in two NNs\footnote{MobileBERT was not used for this experiment, since the inference execution of MobileBERT on co-processors is not supported by any middleware yet.} running on different processors in Mi8Pro. 
The compute- and memory-intensive FC layers exhibit much longer latency when running on co-processors, while other layers exhibit longer latency when running on CPUs. Due to this difference, NNs which have a larger number of FC layers (e.g., MobilenetV3) run more efficiently on CPUs, while others (e.g., InceptionV1) run more efficiently on co-processors. This result also implies that the co-processors do not always outweigh the CPUs, so that carefully choosing one considering layer compositions is crucial for energy efficiency.

\begin{itemize}[leftmargin=*]
\item \textit{The optimal edge-cloud execution varies with the inference quality requirement.}
\end{itemize}

\begin{figure}[t]
    \centering
    \includegraphics[width=\linewidth]{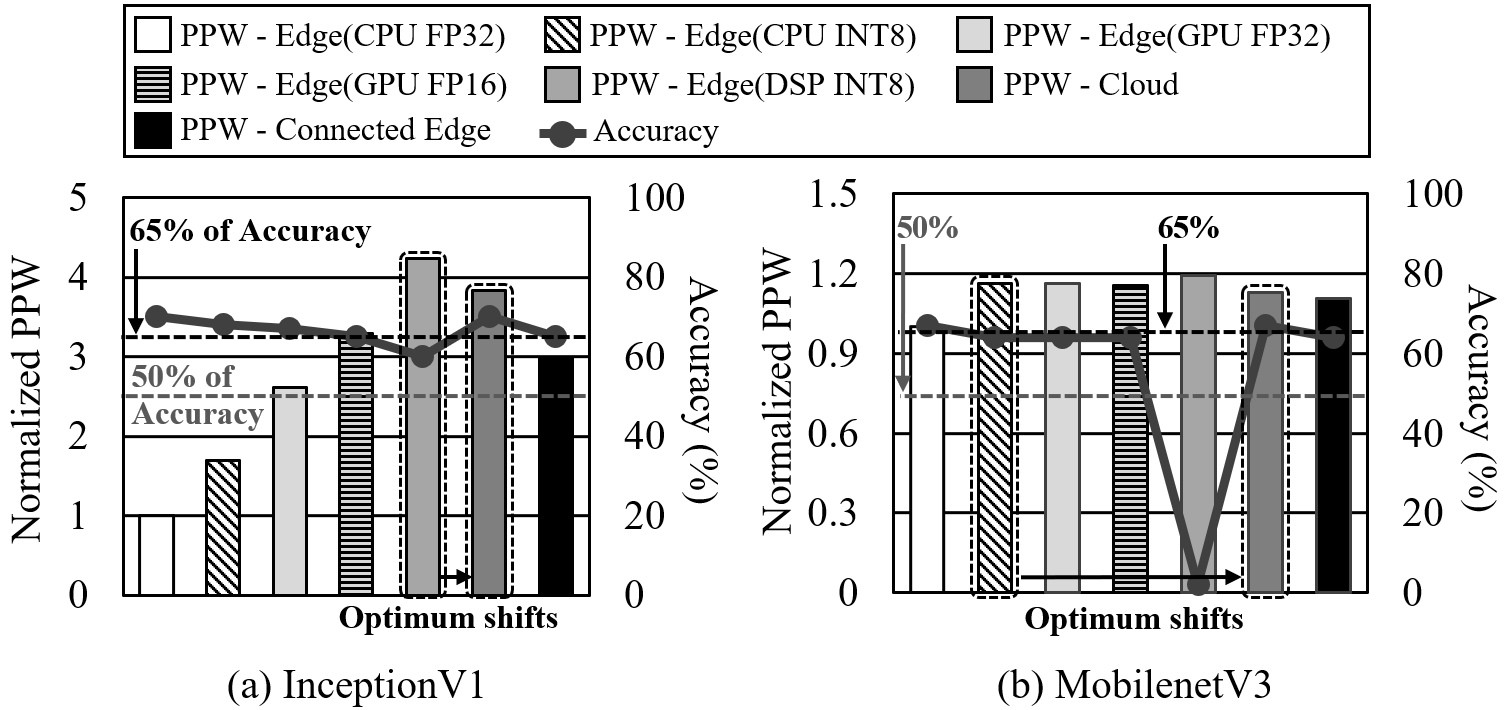}
    \vspace{-0.6cm}
    \caption{Depending on the inference accuracy target, optimal edge-cloud execution also shifts. Note that PPW is normalized to Edge(CPU FP32).}
    \label{fig:energy vs accuracy}
    \vspace{-0.4cm}
\end{figure}

Figure~\ref{fig:energy vs accuracy} shows the energy efficiency (PPW) and accuracy of DNN inference on different execution targets, where the inference quality (i.e., accuracy) of each NN highly depends on the execution target. Note that the accuracy for each processor is measured in our edge-cloud systems, by using ImageNet validation set\cite{JDeng09}.
If the accuracy target is 50\%, the optimal target might be DSP INT8 and CPU INT8 for InceptionV1 and MobilenetV3, respectively; it shows the highest energy efficiency while satisfying the QoS constraint. However, if the accuracy target is 65\%, the optimal target should shift to the cloud to satisfy the accuracy target.

\subsection{Impact of Runtime Variance on Inference Execution}
\label{sec:motivation2}

In a realistic execution environment, there can be on-device interference from co-running applications~\cite{SYLee2017,DShingari2018,YGKim2017_1}. In addition, the network signal strength can significantly vary, depending on the movement of edge device users. In fact, users suffer significant signal strength variations in daily life (43\% of data are transmitted under weak signal strength \cite{NDing2013}).

\begin{itemize}[leftmargin=*]
\item \textit{On-device interference and varying network stability shifts the optimal edge-cloud execution.}
\end{itemize}

\begin{figure}[t]
    \centering
    \includegraphics[width=\linewidth]{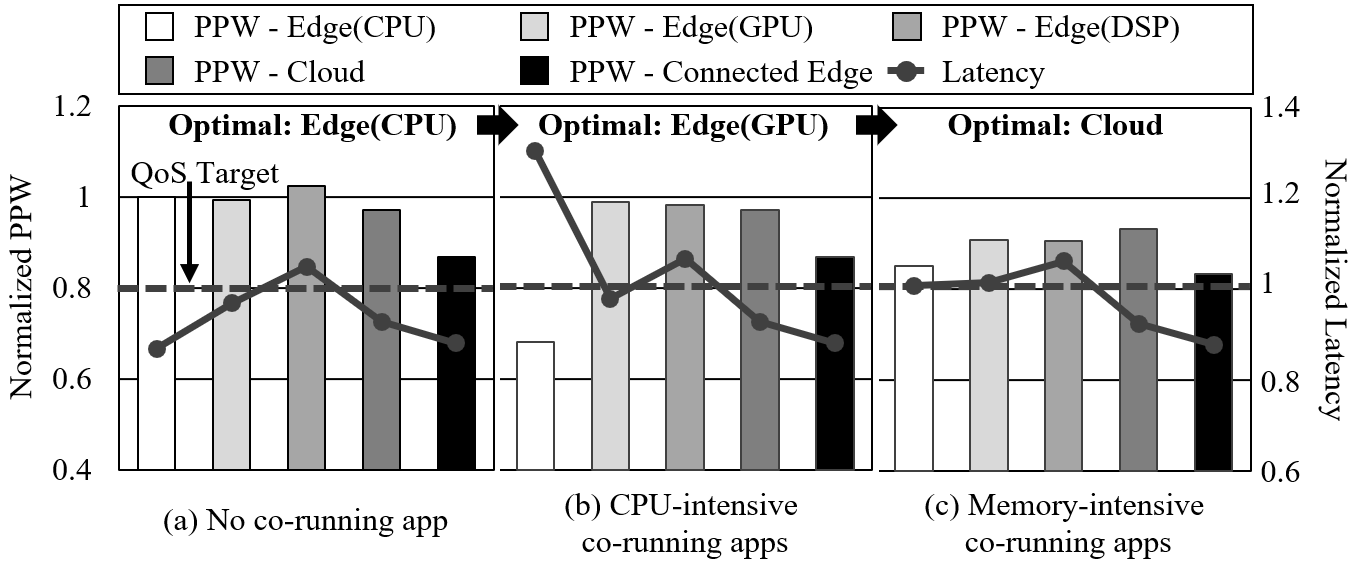}
    \vspace{-0.6cm}
    \caption{In the presence of on-device interference, the optimal edge-cloud execution shifts. Note that PPW is normalized to Edge(CPU) with no co-running app and latency is normalized to the QoS target.}
    \label{fig:interference}
    \vspace{-0.2cm}
\end{figure}

Figure \ref{fig:interference} shows the normalized energy efficiency (PPW) and latency of DNN (i.e., MobilenetV3) inference when CPU-intensive or memory-intensive synthetic applications are co-running, changing the optimal execution target. When a CPU-intensive application is co-running, the energy efficiency of the inference execution on CPU is significantly degraded, due to 1) competitions for CPU resources, and 2) frequent thermal throttling from high CPU utilization~\cite{YGKim2020}. In this case, the optimal execution target shifts from the CPU to the GPU. On the other hand, when a memory-intensive application is co-running, the energy efficiency of all the on-device processors (including CPU, GPU, and DSP) is degraded, since the inference execution is competing with other applications for the memory resources. In this case, the optimal execution target shifts from the edge to the cloud.

\begin{figure}[t]
    \centering
    \includegraphics[width=\linewidth]{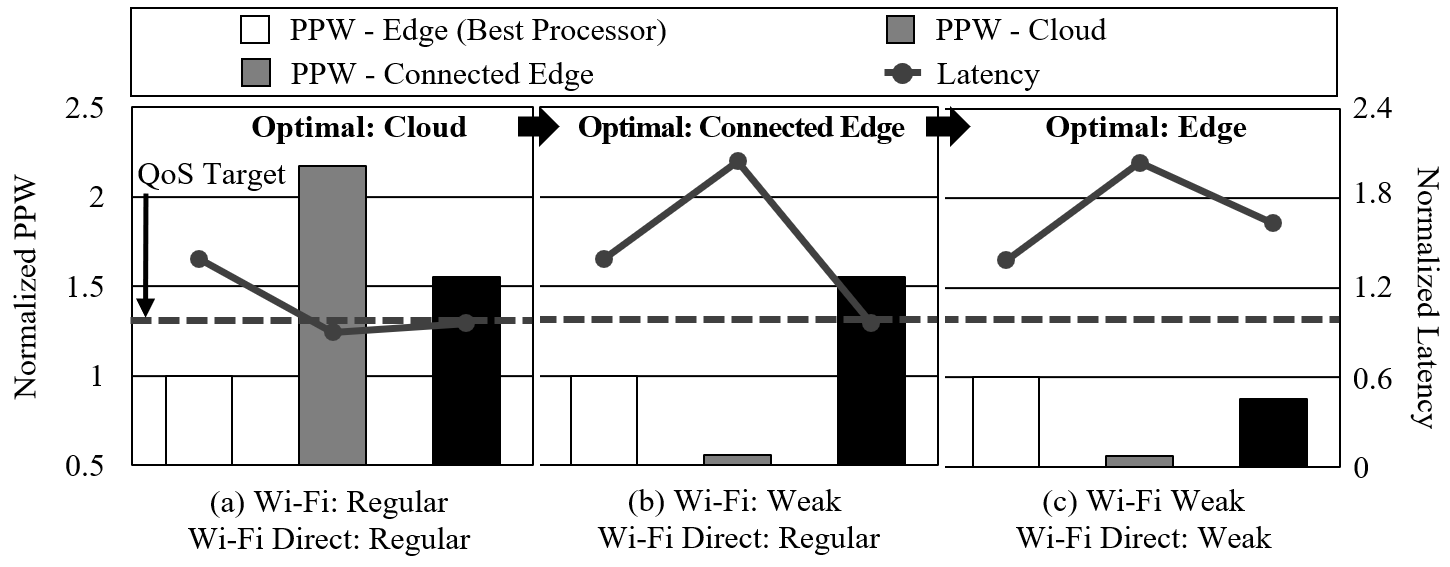}
    \vspace{-0.8cm}
    \caption{Under the signal strength variation, the optimal target for edge-cloud execution also shifts. Note that PPW is normalized to Edge(Best Processor) and latency is normalized to the QoS target.}
    \label{fig:signal strength}
    \vspace{-0.4cm}
\end{figure}

Figure~\ref{fig:signal strength} shows the normalized energy efficiency (PPW) and latency of DNN (i.e., Resnet50) inference when signal strength of wireless networks vary. When the signal strength gets weaker, the energy efficiency of inference execution on the connected systems is significantly degraded, since 1) the data transmission time exponentially increases with decreased data rate~\cite{NDing2013,YGKim2019}, and 2) the network interface consumes more power to transmit data with stronger signals. If only the Wi-Fi signal strength gets weaker, the locally-connected edge device can still serve as an optimal execution target. However, if the signal strength of Wi-Fi direct also gets weak, the optimal target would shift to the edge.

\subsection{Inefficiency of Prediction-based Approaches}
\label{sec:motivation3}

The energy optimization of mobile DNN inference can be formulated as the problem of choosing the optimal execution target under the presence of stochastic runtime variances, which optimizes energy efficiency while satisfying the QoS and accuracy constraints. One of the possible solutions for this kind of problems is to evaluate all the execution targets based on a prediction model. Unfortunately, due to the massive design space, it is difficult to simply build an accurate prediction model. The inaccurate prediction can result in the selection of a sub-optimal execution target.

\begin{itemize}[leftmargin=*]
\item \textit{It is infeasible to enumerate the massive design space exhaustively. Simple prediction-based approaches are insufficient, leaving a significant room for energy efficiency improvement.}
\end{itemize}

To shed light on the inefficiency of existing prediction-based approaches, we compare two types of prediction-based approaches with the baseline (Edge CPU) and oracular design (Opt): (1) regression-based approaches and (2) classification-based approaches. For each type of approaches, we use methods that are widely adopted by existing works in this domain~\cite{AEEshratifar2018,MHan2019,YKang2017,ECai2017,GZhong2019}. For the regression-based approaches, we use Linear Regression (LR)~\cite{GAFSeber2012} and Support Vector Regression (SVR)~\cite{HDrucker1997}. On the other hand, for the classification-based approaches, we use Support Vector Machine (SVM)~\cite{JAKSuykens99} and K-Nearest Neighbor (KNN)~\cite{BZhang04}.

\begin{figure}[t]
    \centering
    \includegraphics[width=\linewidth]{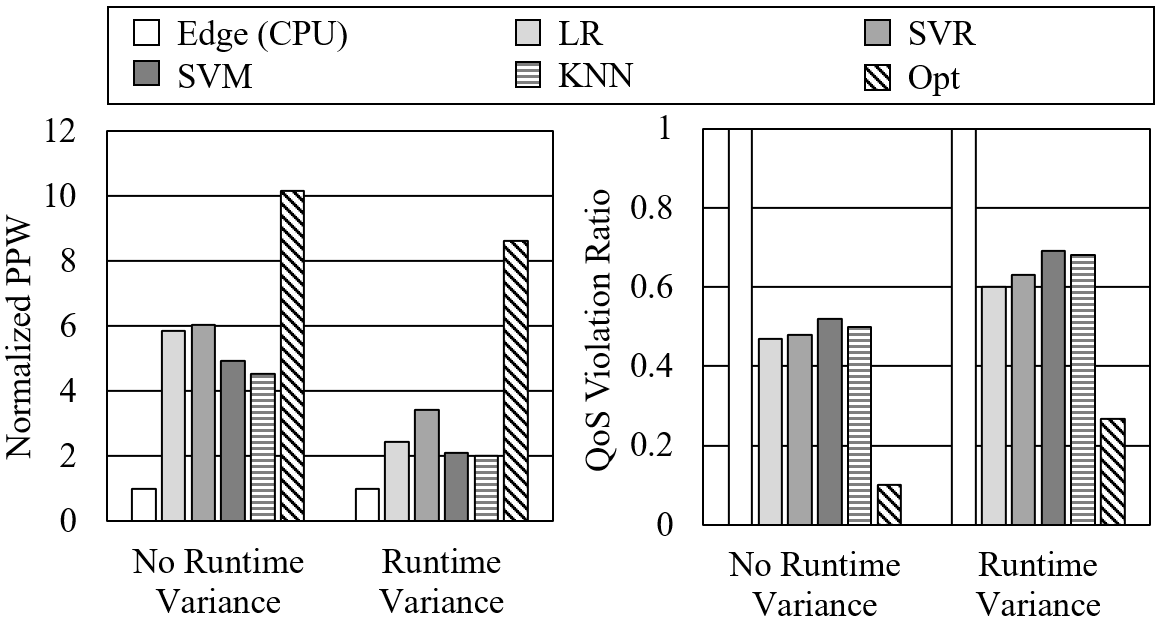}
    \vspace{-0.8cm}
    \caption{There is a significant gap between Opt and existing prediction-based approaches, as they fail to accurately predict the optimal execution target under the presence of runtime variances.}
    \label{fig: prediction-based}
    \vspace{-0.5cm}
\end{figure}

Figure~\ref{fig: prediction-based} shows the energy efficiency (PPW) and the QoS violation ratio of prediction-based approaches normalized to those of Edge CPU. Although the prediction-based approaches improve energy efficiency compared to the baseline, there is a significant gap between the approaches and Opt, as they fail to accurately select the optimal execution target. 

When there is no runtime variance, the MAPEs (Mean Absolute Percentage Errors) of LR and SVR are 13.6\% and 10.8\%, respectively. However, under the presence of stochastic runtime variances, MAPEs of LR and SVR are 24.6\% and 21.1\%, respectively. Due to the inaccurate prediction of energy efficiency and latency, these approaches fail to run DNN inference on the optimal execution target, degrading energy efficiency and violating the QoS constraint.

On the other hand, the miss-classification ratio of SVM and KNN are 12.7\% and 14.3\%, respectively, under the presence of runtime variances. Though the two values do not seem to be large, these approaches degrade energy efficiency and latency much more than regression model-based approaches. This is because the wrong decision occurs regardless of the absolute magnitude of energy efficiency and latency. For example, even though the on-device inference is much more efficient than cloud inference in case of weak signal strength, cloud inference can be selected as the execution target. 

These results call for the need of a novel scheduler design which can accurately select the optimal DNN inference execution target, while adapting to the stochastic runtime variances. In the next section, we explain our proposed \textit{AutoScale} which self-learns the optimal execution target under the presence of runtime variances based on reinforcement learning.

\section{AutoScale}
\label{sec:design}

\begin{figure*}[t]
    \includegraphics[width=\linewidth]{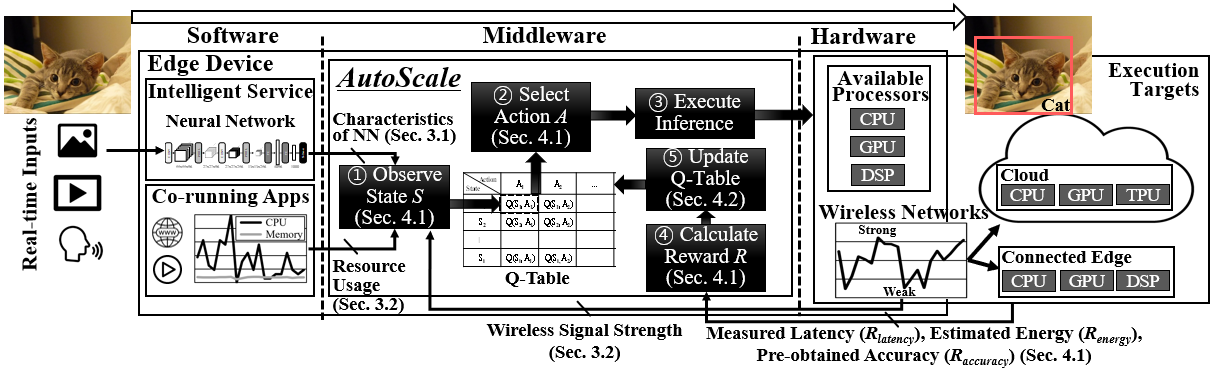} 
    \vspace{-0.8cm}
    \caption{AutoScale design overview.}
    \label{fig:design overview}
    \vspace{-0.5cm}
\end{figure*}

Figure~\ref{fig:design overview} provides the design overview of \textit{AutoScale} in the context of the mobile and edge-cloud DNN inference execution. For each inference execution, \textit{AutoScale} observes the current execution state (\textcircled{\small{1}}), including NN characteristics as well as runtime variances. For the observed state, \textit{AutoScale} selects an action (i.e., execution target) (\textcircled{\small{2}}), which is expected to maximize energy efficiency satisfying QoS and inference quality target, based on a lookup table (i.e., Q-table); the table contains accumulated rewards of the previous selections. \textit{AutoScale} executes DNN inference on the target defined by the selected action (\textcircled{\small{3}}), while observing its result (i.e., energy, latency, and inference accuracy). Based on the observed result, \textit{AutoScale} calculates the reward (\textcircled{\small{4}}), which indicates how much the selected action improves energy efficiency and satisfies QoS and accuracy targets. Finally, \textit{AutoScale} updates Q-table with the calculated reward (\textcircled{\small{5}}).

\textit{AutoScale} leverages Reinforcement Learning (RL) as an adaptive prediction mechanism. Generally, an RL agent learns a policy to select the best action for a given state, based on accumulated rewards \cite{SPagani2020}. In the context of mobile and edge-cloud inference execution, \textit{AutoScale} learns a policy to select the optimal inference execution target for the given NN under the presence of runtime variances, based on the accumulated energy, latency, and accuracy results of selections. To solve system optimization with RL, there are three important design requirements for mobile deployments.

\textbf{High Prediction Accuracy:} The success of RL depends on how much the predicted execution target is close to the optimal one. For the accurate prediction, it is important to correctly model the core components---\textit{State}, \textit{Action}, and \textit{Reward}---in a realistic environment. We define these components based on our observations of a realistic edge inference execution environment (Section~\ref{sec:design1}). 

In addition to the core components, it is also important to avoid local optima. This is deeply related to a classical RL problem, exploitation versus exploration dilemma \cite{EE-Dar2006,DEKoulouriotis2008}. If an RL agent always exploits an action with the temporary highest reward, it might get stuck in local optima. On the other hand, if it keeps exploring all possible actions, the convergence might get slower. To solve this problem, we employ epsilon-greedy algorithm, which is one of the widely adopted randomized greedy algorithms in this domain \cite{SKMandal2020,RNishtala2017,SPagani2020}, due to its simplicity and effectiveness (Section \ref{sec:design2}).

\textbf{Minimal Training Overhead:} In case of RL, training is continuously performed on-device. Due to this reason, reducing training overhead is crucial, particularly for the energy-constrained edge domain. As we observed in Section \ref{sec:motivation}, although performance of execution targets vary across heterogeneous devices, they share similar energy trend for each NN. An RL model trained in a device implicitly has this energy trend knowledge. Hence, we consider transferring a model trained from one device for other devices to expedite the convergence, which might reduce the training overhead. (detailed results are presented in Section \ref{sec:result5}).

\begin{scriptsize}
\begin{table*}[t]
  \centering
  \begin{tabular}{|l|l|l|l|}
    \hline
    \multicolumn{2}{|l|}{\textbf{State}} & \textbf{Description} & \textbf{Discrete values} \\
    \hline
    \hline
        \multirow{4}{1.2cm}{NN-related features} & $S_{CONV}$ & \# of CONV layers & Small (<30), Medium (<50), Large (<90), Larger (>=90) \\ \cline{2-4}
        & $S_{FC}$ & \# of FC layers & Small (<10), Large (>=10) \\ \cline{2-4}
        & $S_{RC}$ & \# of RC layers & Small (<10), Large (>=10) \\  \cline{2-4}
        & $S_{MAC}$ & \# of MAC operations & Small (<1000M), Medium (<2000M), Large (>=2000M) \\  \hline
        \multirow{4}{1.2cm}{Runtime variances} & $S_{Co\_CPU}$ & CPU utilization of 
        co-running apps & None (0\%), Small (<25\%), Medium (<75\%), Large (100\%)\\ \cline{2-4}
        & $S_{Co\_MEM}$ & Memory usage of co-running apps & None (0\%), Small (<25\%), Medium (<75\%), Large (100\%)\\ \cline{2-4}
        & $S_{RSSI\_W}$ & RSSI of wireless local area network & Regular (>-80dBm), Weak (<=-80dBm) \\ \cline{2-4}
        & $S_{RSSI\_P}$ & RSSI of peer-to-peer wireless network & Regular (>-80dBm), Weak (<=-80dBm) \\ \hline
  \end{tabular}
  \vspace{-0.3cm}
  \caption{State-related features}
  \vspace{-0.4cm}
  \label{table:States}
\end{table*}
\end{scriptsize}

\textbf{Low Latency Overhead:} For the real-time inference execution on the energy constrained edge devices, latency overhead is also one of the crucial factors. Among the various forms of RL \cite{SPagani2020}, such as Q-learning~\cite{YChoi2019}, TD-learning~\cite{XLin2016}, and deep RL~\cite{VMnih2015}, Q-learning has an advantage for low latency overhead, as it finds the best action with a look-up table. Hence, in this paper, we use Q-learning for \textit{AutoScale}.

\subsection{AutoScale RL Design}
\label{sec:design1}

In RL, there are three core components: {\it State}, {\it Action}, and {\it Reward}. In this section, we define the core components to formulate the optimization space for \textit{AutoScale}.

\textbf{State -} Based on the observations presented in Section \ref{sec:motivation}, we identify states that are critical to edge inference execution. Table~\ref{table:States} summarizes the states.

As we explored in Section \ref{sec:motivation1}, the optimal execution target depends on layer compositions of NNs. However, identifying states with all the layer types is not desirable, since the latency overhead (i.e., Q-table lookup time) increases. Hence, we identify states with layer types that are deeply correlated to the energy efficiency and performance of inference execution. We test the correlation strength between each layer type and energy/latency by calculating the squared correlation coefficient ($\rho^{2}$) \cite{YZhu2013}. We find CONV, FC, and RC layers are the most correlated to the energy efficiency and performance, due to their compute- and/or memory-intensive natures. Thus, we identify $S_{CONV}$, $S_{FC}$, and $S_{RC}$ which represent the number of CONV, FC, and RC layers in NNs, respectively. We also identify $S_{MAC}$, the number of MAC operations to consider heaviness of NNs.

As we explored in Section \ref{sec:motivation2}, the efficiency of edge inference highly depends on the CPU-intensity and memory-intensity of co-running applications. Hence, we use $S_{Co\_CPU}$ and $S_{Co\_MEM}$ which represent the CPU utilization and memory usage of co-running applications, respectively. In addition, the efficiency of inference execution on the connected systems highly depends on the signal strength of wireless networks. For this reason, we use $S_{RSSI\_W}$ and $S_{RSSI\_P}$ which stand for the RSSI of wireless local area network (e.g., Wi-Fi, LTE, and 5G) and RSSI of peer-to-peer wireless network (e.g., Bluetooth, Wi-Fi direct, etc.), respectively. 

When a feature has a continuous value, it is difficult to define the state in a discrete manner for the lookup table of Q-learning \cite{YChoi2019, RNishtala2017}. To convert the continuous features into discrete values, we applied DBSCAN clustering algorithm to each feature \cite{YChoi2019}; DBSCAN determines the optimal number of clusters for the given data. The last column of Table \ref{table:States} summarizes discrete values for each state. 

\textbf{Action -} Actions in reinforcement learning represent the choosable control knobs of the system. In the context of the edge-cloud inference execution, we define the actions as the available execution targets. For the edge inference execution, available processors in mobile SoCs, such as CPUs, GPUs, DSPs, and NPUS, are defined as the actions. On the other hand, for the cloud execution, server-class processors, such as CPUs, GPUs, and TPUs, are defined as the actions. 

The set of actions can be augmented to consider other control knobs, such as Dynamic Voltage and Frequency Scaling (DVFS) and quantization. For example, as long as the QoS constraint is satisfied, it is possible to reduce the frequency of processors, saving energy. In addition, employing the quantization for each processor can reduce both compute and memory intensities of the inference execution, improving energy efficiency and performance.

\textbf{Reward -} Reward in RL models the optimization objective of the system. To represent the three important optimization axes, we encode three different rewards, $R_{latency}$, $R_{energy}$, and $R_{accuracy}$. $R_{latency}$ is the measured inference latency for a selected action (i.e., execution target for DNN inference). On the other hand, $R_{energy}$ is the estimated energy consumption of the selected action. $R_{accuracy}$ is pre-measured inference accuracy of the given NN on each execution target.

We estimate $R_{energy}$ of edge execution as follows. When the CPU is selected as the action, $R_{energy}$ is calculated using the utilization-based CPU power model \cite{YKim2019,LZhang2010} as in (\ref{eq: CPU energy}), where $E^{i}_{Core}$ is the power consumed by the i-th core, $t^{f}_{busy}$ and $t_{idle}$ are the time spent in the busy state at frequency \textit{f} and that in the idle state, respectively, and $P^{f}_{busy}$ and $P_{idle}$ are power consumed during $t^{f}_{busy}$ at \textit{f} and that during $t_{idle}$, respectively.
\begin{equation}
    \label{eq: CPU energy}
    \vspace{-0.1cm}
    \begin{aligned}
        R_{energy} &= \sum_{i}{E_{Core}^{i}}, \\
        E_{Core} &= \sum_{f}(P_{busy}^{f} \times t_{busy}^{f}) + P_{idle} \times t_{idle} \\
    \end{aligned}
    \vspace{-0.2cm}
\end{equation}
Similarly, if scaling out the inference execution to GPUs within the system is selected as the action, $R_{energy}$ is is calculated using the utilization-based GPU power model \cite{YGKIM2015} as in (\ref{eq: GPU energy}). Note $P^{f}_{busy}$ and $P_{idle}$ values for CPU/GPU are obtained from \textit{procfs} and \textit{sysfs} in Linux kernel \cite{YGKim2017_1}, while $P^{f}_{busy}$ and $P_{idle}$ values for CPU/GPU are obtained by measuring power consumption of CPU/GPU at each frequency in the busy state and that in the idle state, respectively, and stored in a look-up table of {\em AutoScale}.
\begin{equation}
    \vspace{-0.1cm}
    \label{eq: GPU energy}
    R_{energy} = \sum_{f}(P_{busy}^{f} \times t_{busy}^{f}) + P_{idle} \times t_{idle}
    \vspace{-0.1cm}
\end{equation}
If scaling out the inference execution to DSPs is selected as the action, $R_{energy}$ is calculated as in (\ref{eq: DSP energy}), where $P_{DSP}$ is a pre-measured power consumption of DSP; we use the constant value for $P_{DSP}$, since $P_{DSP}$ was consistent during 100 inference runs of 10 NNs. 
\vspace{-0.1cm}
\begin{equation}
    \label{eq: DSP energy}
    E_{DSP} = P_{DSP} \times R_{latency}
    \vspace{-0.1cm}
\end{equation}
On the other hand, if scaling out the inference execution to connected systems is selected as the action, $R_{energy}$ is calculated using the signal strength-based energy model \cite{YGKim2019} as in (\ref{eq: offloading}), where $t_{TX}$ and $t_{RX}$ are data transmission latency measured while transmitting the input and receiving the output, respectively and $P^{S}_{TX}$ and $P^{S}_{RX}$ are power consumed by a wireless network interface during $t_{TX}$ and $t_{RX}$ at signal strength \textit{S}, respectively. Note $P^{S}_{TX}$ and $P^{S}_{RX}$ values for each network are obtained by measuring power consumption of wireless network interfaces at each signal strength while transmitting and receiving data, respectively.
\begin{equation}
    \label{eq: offloading}
    \begin{aligned}
        R_{energy} = &P_{TX}^{S} \times t_{TX} \;\; + \;\; P_{RX}^{S} \times t_{RX} \\
        &+ \;\; P_{idle} \times (R_{latency} - t_{TX} - t_{RX})
    \end{aligned}
\end{equation}
Since the energy estimation is based on the measured latency, the Mean Absolute Percentage Error (MAPE) of the energy estimation is 7.3\%, which is low enough to identify the optimal action.

To make {\em AutoScale} learn and select an efficient execution decision which maximizes energy efficiency satisfying the QoS and accuracy constraints, the reward \textit{R} is calculated as in (\ref{eq: reward}), where $\alpha$ and $\beta$ are the weights of latency and accuracy, respectively; we use 0.1 for both weights, but we can use higher weights if the inference workload requires higher performance and accuracy.
\begin{equation}
    \label{eq: reward}
    \begin{aligned}
        &if \;\;\; R_{accuracy} \; < \;\; Inference \;\; Quality \;\; Requirement,\\
        &\qquad \qquad R = - R_{accuracy}\\
        &else \\
        &\qquad if \;\;\; R_{latency} \; < \;\; QoS \;\; Constraint,\\
            &\qquad \qquad R = - R_{energy} + \alpha R_{latency} + \beta R_{accuracy}\\
        &\qquad else\\
            &\qquad \qquad R = - R_{energy} + \beta R_{accuracy}\\
    \end{aligned}
\end{equation}
If the inference quality requirement of the selected action is not satisfied, $R_{accuracy}$ multiplied by -1 is used as the reward value, to avoid choosing the target from the next inference running. Otherwise, the reward value is calculated depending on whether the QoS constraint is satisfied or not. In (\ref{eq: reward}), $R_{energy}$ is multiplied by -1, to produce higher rewards for lower energy consumption. 

\subsection{AutoScale Implementation}
\label{sec:design2}

As previously discussed, we use Q-learning for \textit{AutoScale}'s implementation due to its low runtime overhead. To deal with the exploitation versus exploration dilemma in RL, we also employ the epsilon-greedy algorithm for \textit{AutoScale}, which chooses the action with the highest reward or a uniformly random action based on an exploration probability. 

\begin{algorithm}[t]
\caption{Training Q-Learning Model}
\label{alg: training}
\textbf{Variable}: \textit{S, A} \\
\hspace*{\algorithmicindent}\hspace*{\algorithmicindent} \textit{S} is the variable for the state \\
\hspace*{\algorithmicindent}\hspace*{\algorithmicindent} \textit{A} is the variable for the action \\
\textbf{Constants}: \textit{$\gamma$, $\mu$, $\epsilon$} \\
\hspace*{\algorithmicindent}\hspace*{\algorithmicindent} \textit{$\gamma$} is the learning rate \\
\hspace*{\algorithmicindent}\hspace*{\algorithmicindent} \textit{$\mu$} is the discount factor \\
\hspace*{\algorithmicindent}\hspace*{\algorithmicindent} \textit{$\epsilon$} is the exploration probability \\
\textbf{Initialize} \textit{Q(S,A)} as random values \\
\textbf{Repeat} (whenever inference starts): \\
\hspace*{\algorithmicindent}\hspace*{\algorithmicindent} Observe state and store in \textit{S} \\
\hspace*{\algorithmicindent}\hspace*{\algorithmicindent} \textbf{if} \textit{rand()} < \textit{$\epsilon$} \textbf{then} \\
\hspace*{\algorithmicindent}\hspace*{\algorithmicindent}\hspace*{\algorithmicindent} Choose action \textit{A} randomly \\
\hspace*{\algorithmicindent}\hspace*{\algorithmicindent}\textbf{else} \\
\hspace*{\algorithmicindent}\hspace*{\algorithmicindent}\hspace*{\algorithmicindent} Choose action \textit{A} which maximizes \textit{Q(S,A)} \\
\hspace*{\algorithmicindent}\hspace*{\algorithmicindent} Run inference on a target defined by \textit{A}\\
\hspace*{\algorithmicindent}\hspace*{\algorithmicindent} (when inference ends) \\
\hspace*{\algorithmicindent}\hspace*{\algorithmicindent} Measure $R_{latency}$, estimate $R_{energy}$, and obtain $R_{accuracy}$ \\
\hspace*{\algorithmicindent}\hspace*{\algorithmicindent} Calculate reward \textit{R} \\
\hspace*{\algorithmicindent}\hspace*{\algorithmicindent} Observe new state \textit{S'} \\
\hspace*{\algorithmicindent}\hspace*{\algorithmicindent} Choose action \textit{A'} which maximizes \textit{Q(S',A')} \\
\hspace*{\algorithmicindent}\hspace*{\algorithmicindent} \textit{Q(S,A)} $\leftarrow$ \textit{Q(S,A)} +  $\gamma$[\textit{R} + $\mu$\textit{Q(S',A')} - \textit{Q(S,A)}] \\
\hspace*{\algorithmicindent}\hspace*{\algorithmicindent} \textit{S} $\leftarrow$ \textit{S'} 
\end{algorithm}

In Q-learning, the value function, denoted as \textit{Q(S,A)}, takes State \textit{S} and Action \textit{A} as parameters. \textit{Q(S,A)} is a form of a look-up table, called Q-table. Algorithm \ref{alg: training} shows the detailed algorithm for training the Q-table for on-device DNN inference. At the beginning, the Q-table is initialized with random values. At runtime, for each DNN inference, the algorithm observes \textit{S} by checking the NN characteristics and runtime variances. For the given \textit{S}, the algorithm evaluates a random value compared to $\epsilon$\footnote{Note we use 0.1 for the $\epsilon$ value by referring to previous RL-based works in this domain \cite{SKMandal2020,RNishtala2017}.}. If the random value is smaller than $\epsilon$, the algorithm randomly chooses \textit{A} for exploration. 
Otherwise, the algorithm chooses \textit{A} with the largest \textit{Q(S,A)}. After choosing \textit{A}, the algorithm runs the inference on a target defined by \textit{A}. During the inference, the algorithm measures $R_{latency}$ and estimates $R_{energy}$, as explained in Section \ref{sec:design1}. In addition, it obtains $R_{accuracy}$ from the stored inference accuracy of the given NN on the selected execution target. Based on these values, the algorithm calculates reward \textit{R} as in (\ref{eq: reward}) of Section \ref{sec:design1}. After calculating the \textit{R} value, the algorithm observes new state \textit{S'} and chooses \textit{A'} for the given \textit{S'} with the largest \textit{Q(S',A')}. The algorithm updates the \textit{Q(S,A)} based on the equation in Algorithm \ref{alg: training}. In the equation for updating the \textit{Q(S,A)}, $\gamma$ and $\mu$ are hyperparameters, which represent the learning rate and the discount factor, respectively. The learning rate indicates how much the newly acquired information overrides the old information. On the other hand, the discount factor gives more weight to the rewards in the near future. We set $\gamma$ and $\mu$, based on a sensitivity test on hyperparameters (details are explained in Section \ref{sec:methodology3}).

After the learning is completed (i.e., the largest \textit{Q(S,A)} value for each state \textit{S} is converged), the Q-table is used to select \textit{A} which maximizes \textit{Q(S,A)} for the observed \textit{S}.
\section{Experimental Methodology}
\label{sec:methodology}

\subsection{Real System Measurement Infrastructure}
\label{sec:methodology1}

We perform our experiments on three smartphones---Mi8Pro \qquad ~\cite{Huawei}, Galaxy S10e~\cite{Samsung}, and Moto X Force~\cite{Motorola}. Table~\ref{table:Devices} summarizes their specifications\footnote{Though there exist lower-performance cores in mobile CPUs, we only present the high-performance cores, since DNN inference usually run on the high-performance cores.}.
Note we only use the smartphone with DSP rather than that with NPU, since 1) NPUs are only programmable with vendor-provided Software Development Kits (SDKs) which have not been publicly released yet~\cite{AIgnatov2018}, and 2) DSPs in recent mobile SoCs are optimized for DNN inference so that they can act as NPUs~\cite{AIgnatov2018, Qualcomm}.

For cloud inference execution, we connect the smartphones to a server, equipped with an Intel Xeon CPU E5-2640 with 2.4GHz of 40 cores, NVIDIA Tesla P100 GPU, and 256 GB of RAM, via Wi-Fi. 
To control the Wi-Fi signal strength, we adjust the distance between the smartphones and Wi-Fi Access Point (AP).
For inference execution on locally connected edge, we use a tablet, Galaxy Tab S6, equipped with 2.84GHz of Cortex A76 CPU, Adreno 640 GPU, and Hexagon 690 DSP. We connect the smartphones to the tablet through Wi-Fi direct, one of the Wi-Fi-based peer-to-peer wireless networks. To control the signal strength of Wi-Fi Direct, we adjust the distance between the locally connected devices.
We measure the system-wide power consumption of the smartphones using an external Monsoon Power Meter~\cite{Monsoon} -- similar practice is used in a number of prior works~\cite{Pandiyan:IISWC14,Carroll:Usenix10,Bircher:ISPASS07}.

\begin{scriptsize}
\begin{table}[t]
  \centering
  \begin{tabular}{|l|l|l|l|}
    \hline
    \textbf{Device}& \textbf{CPU}& \textbf{GPU}& \textbf{DSP} \\
    \hline
    \hline
    \multirow{4}{1.2cm}{Mi8Pro} & Cortex A75 -& \multirow{4}{1.8cm}{Adreno 630 - 0.7GHz w/ \qquad 7 V/F steps (2.8 W)}& \multirow{4}{1.8cm}{Hexagon 685 (1.8 W)}\\ 
    & 2.8GHz w/& &\\
    & 23 V/F steps & & \\
    & (5.5 W) & & \\
    \hline
    \multirow{4}{1.2cm}{Galaxy S10e} & Mongoose -& \multirow{4}{1.8cm}{Mali-G76 - 0.7GHz w/ \qquad 9 V/F steps (2.4 W)}& \multirow{4}{1.5cm}{-}\\ 
    & 2.7GHz w/ & &\\
    & 21 V/F steps &  & \\
    & (5.6 W) & &  \\
    \hline
    \multirow{4}{1.2cm}{Moto X Force} & Cortex A57 -& \multirow{4}{1.8cm}{Adreno 430 - 0.6GHz w/ \qquad 6 V/F steps (2.0 W)}& \multirow{4}{1.5cm}{-}\\ 
    & 1.9GHz w/ & &\\
    & 15 V/F steps & & \\
    & (3.6 W) & &\\
    \hline
  \end{tabular}
    \vspace{-0.2cm}
  \caption{Mobile device specification with the peak power consumption shown in the parenthesis.}
  \vspace{-0.2cm}
  \label{table:Devices}
\end{table}
\end{scriptsize}

\begin{scriptsize}
\begin{table}[t]
  \centering
  \begin{tabular}{|l|l|l|l|l|}
    \hline
    \textbf{Workload} & \textbf{DNNs} & \textbf{$S_{CONV}$} & \textbf{$S_{FC}$} & \textbf{$S_{RC}$} \\
    \hline
    \hline
    \multirow{6}{1.7cm}{Image classification} & InceptionV1 & 49 & 1 & 0 \\ \cline{2-5} 
        & InceptionV3 & 94 & 1 & 0 \\ \cline{2-5}
        & MobilenetV1 & 14 & 1 & 0 \\ \cline{2-5}
        & MobilenetV2 & 35 & 1 & 0 \\ \cline{2-5} 
        & MobilenetV3 & 23 & 20 & 0 \\ \cline{2-5}
        & Resnet50 & 53 & 1 & 0 \\ \hline
        \multirow{3}{1.7cm}{Object detection} & SSD MobilenetV1 & 19 & 1 & 0 \\ \cline{2-5}
        & SSD MobilenetV2 & 52 & 1 & 0 \\ \cline{2-5}
        & SSD MobilenetV3 & 28 & 20 & 0 \\ \hline
        Translation & MobileBERT & 0 & 1 & 24 \\ \hline
  \end{tabular}
    \vspace{-0.2cm}
  \caption{DNN inference workloads. Layer compositions are obtained from the TensorFlow NN implementations.}
    \vspace{-0.5cm}
  \label{table:DNN Workloads}
\end{table}
\end{scriptsize}

To execute DNN inference on diverse processors in edge-cloud systems, we build on top of TVM~\cite{TChen2018} and SNPE~\cite{Qualcomm}. 
TVM compiles NNs from TensorFlow/TFLite and generates executables for edge/cloud CPUs and GPUs, whereas SNPE complies NNs and generates executables for mobile DSPs. The executables are deployed onto each device with library implementations and are used for edge inference at runtime.

To evaluate the effectiveness of \textit{AutoScale},
we compare \textit{AutoScale} to five baselines available in our edge-cloud systems---(1) Edge(CPU FP32) which always runs DNN inference on the CPU of the edge device, (2) Edge (Best) which runs the inference on the most energy-efficient processor of the edge device, (3) Cloud which always runs inference on the cloud, (4) Connected Edge which always runs inference on another locally connected edge, and (5) Opt, an oracular design which always runs inference on the optimal execution target.

\subsection{Benchmarks and Execution Scenarios}
\label{sec:methodology2}

For our evaluation, we use 10 neural networks that are widely used in real use case scenarios~\cite{MHan2019,VJReddi2019}.
As summarized in Table \ref{table:DNN Workloads}, each NN has different layer compositions.

To explore real use cases, we implement an Android application.
For computer vision workloads (i.e., image classification and object detection), we implement two use case scenarios: non-streaming and streaming. For the non-streaming scenario, the Android application takes an image from the camera and performs inference on the image. For this scenario, short response time is important to users. Since users cannot perceive any notable difference as long as the response time is less than 50 ms \cite{YEndo1996, DLo2015, YZhu2015}, we use 50 ms as the QoS target. On the other hand, for the streaming scenario, the Android application takes a real-time video from the camera and performs inference on the video. For this scenario, high Frames Per Second (FPS) is important for user satisfaction. Since users cannot perceive any difference on the QoS as long as the FPS is greater than 30\cite{BEgilmez2017,YZhu2015}, we consider 30 FPS as the QoS target. For MobileBERT in NLP, we implement one use case scenario, where the Android application performs the translation on a sentence typed by the keyboard. For this scenario, we use 100 ms as the QoS target~\cite{VJReddi2019}.

To validate the effectiveness of \textit{AutoScale} in real execution environment with varying runtime variances, we run our experiments in two execution environments---static and dynamic.
For the static environment, we perform experiments in fixed runtime variances (e.g., co-running apps with constant CPU and memory usages and constant Wi-Fi and Wi-Fi Direct signal strengths).
On the other hand, for the dynamic environment, we perform experiments with varying runtime variances.
To mimic real execution environment, for the co-running app, we implement synthetic applications based on the CPU and memory usage trace of two real-world applications---a web browser and a music player.
In addition, since the signal strength variance is usually modeled with Gaussian distribution~\cite{NDing2013}, we emulate random signal strength with a Gaussian distribution by adjusting the bandwidth limit of the Wi-Fi AP.
Table \ref{table:Execution Environment} summarizes the DNN inference execution environments.

\subsection{AutoScale Design Specification}
\label{sec:methodology3}

\textbf{Actions} We determine actions of \textit{AutoScale} with processors available in our edge-cloud system. Since the energy efficiency of mobile CPU/GPU can be further optimized via DVFS, we identify each Voltage/Frequency (V/F) step of mobile CPU/GPU as the augmented action; the number of available V/F steps is presented in Table \ref{table:Devices}. We do not consider DVFS for DSP in our experiments, since DSP does not support DVFS yet. We also identify the quantization available for each mobile processor (INT8 for CPU and DSP, and FP16 for GPU) as the augmented action.

\textbf{Hyperparameters} To determine two hyperparameters---the learning rate and the discount factor---we evaluate three values of 0.1, 0.5, and 0.9 for each hyperparameter. We observe that higher learning rate is better, which means the more reward is reflected to the Q values, the better \textit{AutoScale} works. We also observe that a lower discount factor is better. This means that the consecutive states have a weak relationship due to the stochastic nature, therefore giving less weight to the rewards in the near future improves the efficiency of \textit{AutoScale}. Thus, in our evaluation, we use 0.9 and 0.1 for the learning rate and discount factor, respectively.

{\bf Training -} To cover the design space of \textit{AutoScale} with sufficient training samples, we repeatedly execute inference 100 times for each NN in each runtime variance-related state (i.e., $S_{CO\_CPU}$, $S_{CO\_MEM}$, $S_{RSSI\_W}$, and $S_{RSSI\_P}$ in Table \ref{table:States}). This results in a total of 64,000 training samples for \textit{AutoScale}. We analyze the training overhead in Section \ref{sec:result5}.

\begin{scriptsize}
\begin{table}[t]
  \centering
  \begin{tabular}{|l|l|l|}
    \hline
    \multicolumn{2}{|l|}{\textbf{Environment}} & \textbf{Description} \\
    \hline
    \hline
    \multirow{5}{*}{Static} & S1 & No runtime variance \\ \cline{2-3} 
    & S2 & CPU-intensive co-running app \\ \cline{2-3}
    & S3 & Memory-intensive co-running app \\ \cline{2-3}
    & S4 & Weak Wi-Fi signal strength \\ \cline{2-3}
    & S5 & Weak Wi-Fi direct signal strength\\ \hline
    \multirow{3}{*}{Dynamic} & D1 & Co-running app - music player \\ \cline{2-3}
    & D2 & Co-running app - web browser \\ \cline{2-3}
    & D3 & Random Wi-Fi signal strength \\ \hline
  \end{tabular}
    \vspace{-0.2cm}
  \caption{DNN inference execution environment}
    \vspace{-0.4cm}
  \label{table:Execution Environment}
\end{table}
\end{scriptsize}
\section {Evaluation Results and Analysis}
\label{sec:result}

\subsection{Performance and Energy Efficiency}
\label{sec:result1}

\begin{figure}[t]
    \centering
    \includegraphics[width=\linewidth]{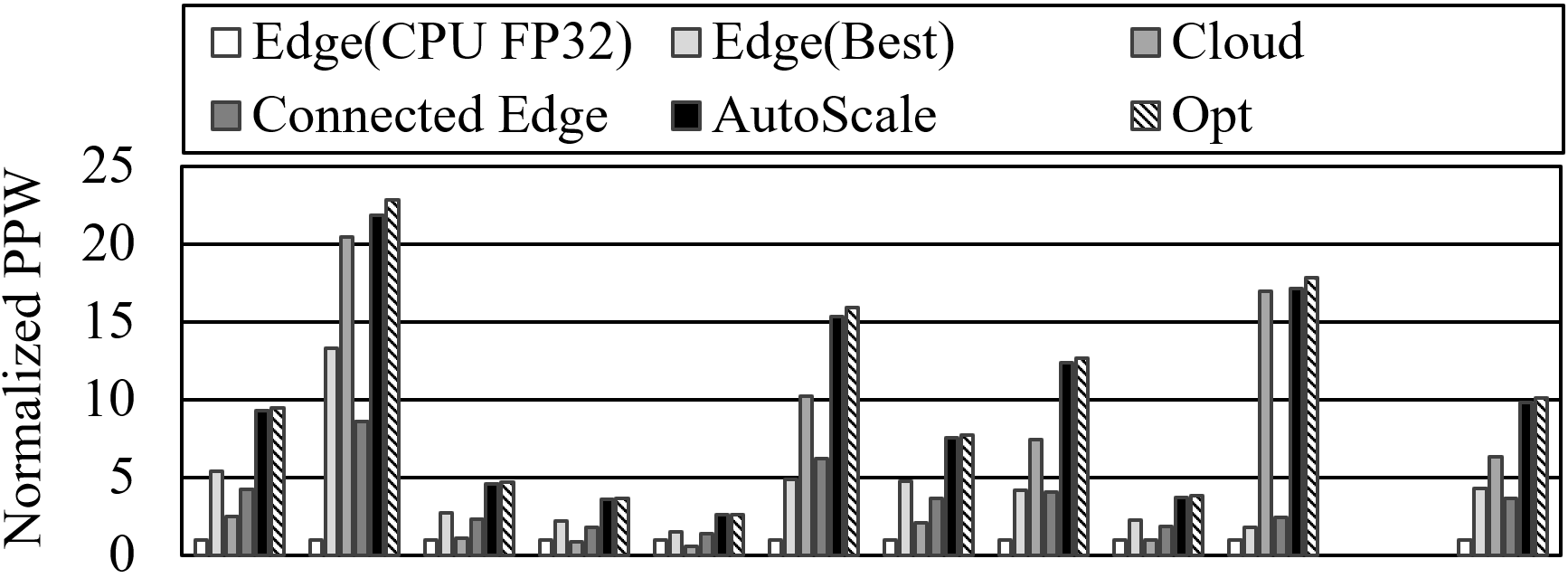}
    \includegraphics[width=\linewidth]{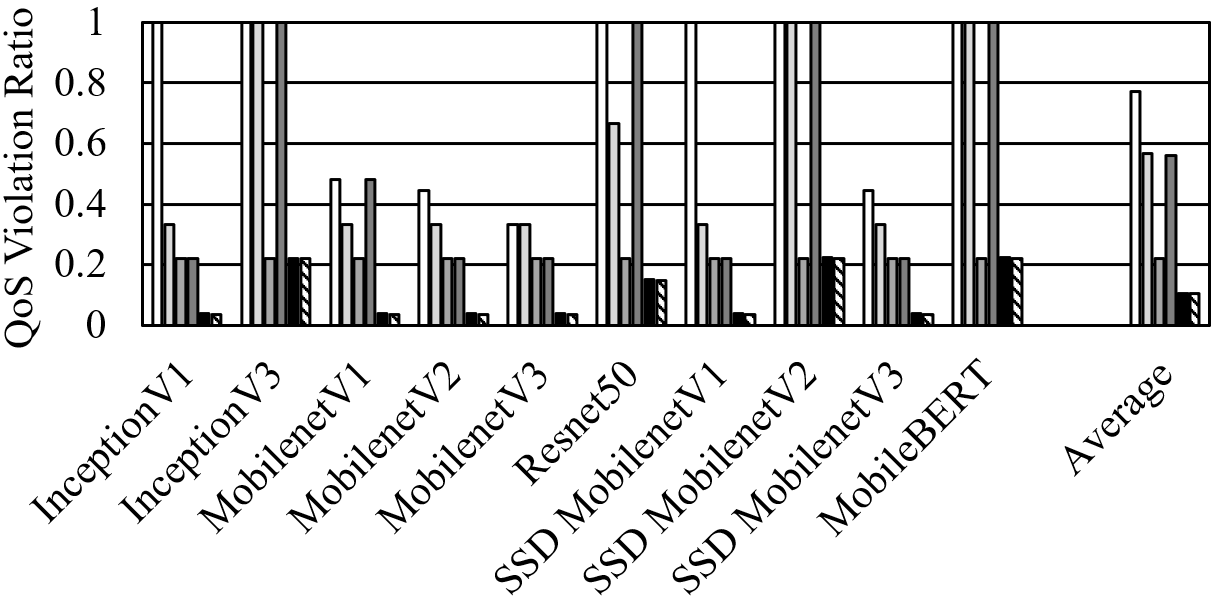}
    \vspace{-0.6cm}
    \caption{\textit{AutoScale} significantly improves energy efficiency compared to baselines satisfying QoS constraints as much as possible.}
    \label{fig:overall}
    \vspace{-0.3cm}
\end{figure}

Figure \ref{fig:overall} shows the average energy efficiency (PPW) normalized to Edge (CPU FP32) and the QoS violation ratio of DNN inference on three mobile devices at static environments. Overall, \textit{AutoScale} improves the average energy efficiency of the DNN inference by 9.8X, 2.3X, 1.6X, and 2.7X, compared to Edge(CPU FP32), Edge(Best), Cloud, and Connected Edge, respectively. Across the diverse collection of neural networks, \textit{AutoScale} can predict the optimal execution target to optimize the energy efficiency of DNN inference, satisfying the QoS constraint as much as possible. \textit{AutoScale} achieves almost the same energy efficiency improvement as Opt; the energy efficiency difference between \textit{AutoScale} and Opt is only 3.2\%.

\begin{figure}[t]
    \includegraphics[width=\linewidth]{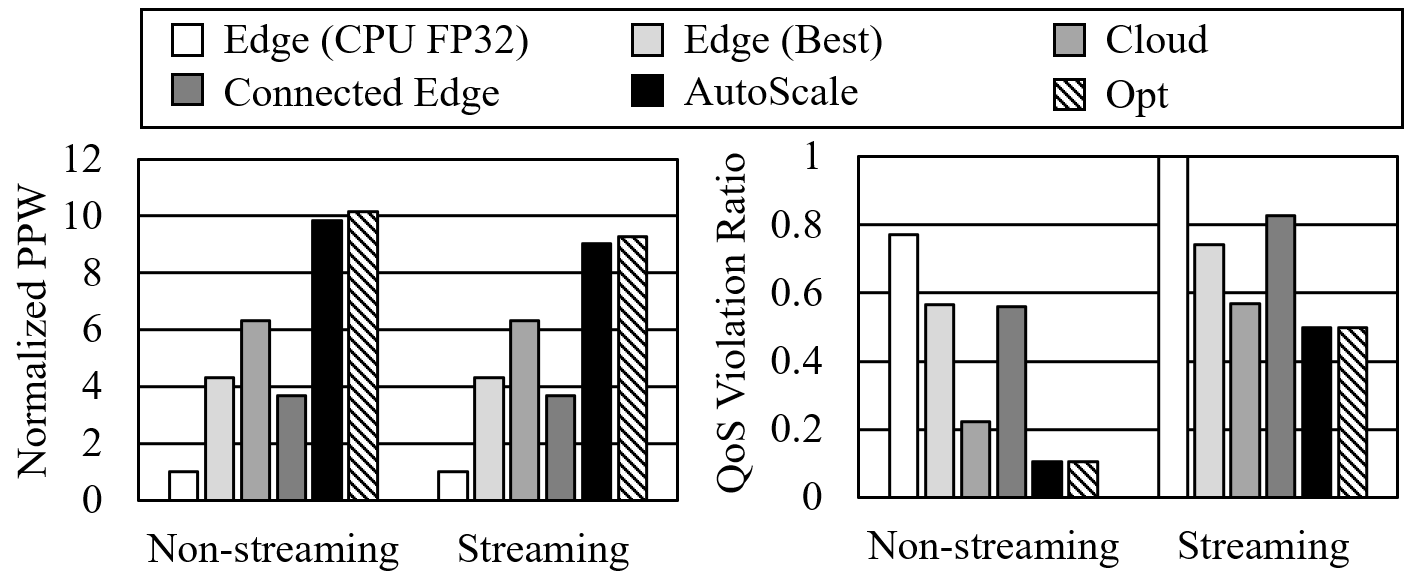}
    \vspace{-0.6cm}
    \caption{Even when the inference intensity increases (i.e., from non-streaming to streaming) \textit{AutoScale} still improves energy efficiency substantially and shows much lower QoS violation ratio, compared to baselines.}
    \label{fig:inference intensity}
    \vspace{-0.2cm}
\end{figure}

In addition, \textit{AutoScale} shows significantly lower QoS violation ratio, compared to the baselines. In fact, \textit{AutoScale} achieves almost the same QoS violation ratio with Opt; the QoS violation ratio difference between \textit{AutoScale} and Opt is only 1.9\%. For light NNs, \textit{AutoScale} does not violate the QoS constraint except for the case when CPU-intensive and memory-intensive applications are co-running or the signal strength of wireless networks is weak. For heavy NNs, \textit{AutoScale} mostly rely on cloud execution so that QoS violation occurs when the signal strength of Wi-Fi is weak. 

When the inference intensity increases (i.e., streaming scenario), the energy efficiency and QoS violation ratio of \textit{AutoScale} is degraded, as shown in Figure \ref{fig:inference intensity}. Nevertheless, \textit{AutoScale} still significantly improves energy efficiency and shows much lower QoS violation ratio, compared to the baselines. In addition, since \textit{AutoScale} accurately selects the optimal execution target regardless of the inference intensity, it achieves almost the same energy efficiency and QoS violation ratio as Opt.

\subsection{Adaptability and Accuracy Analysis}
\label{sec:result2}

\begin{figure}[t]
    \includegraphics[width=\linewidth]{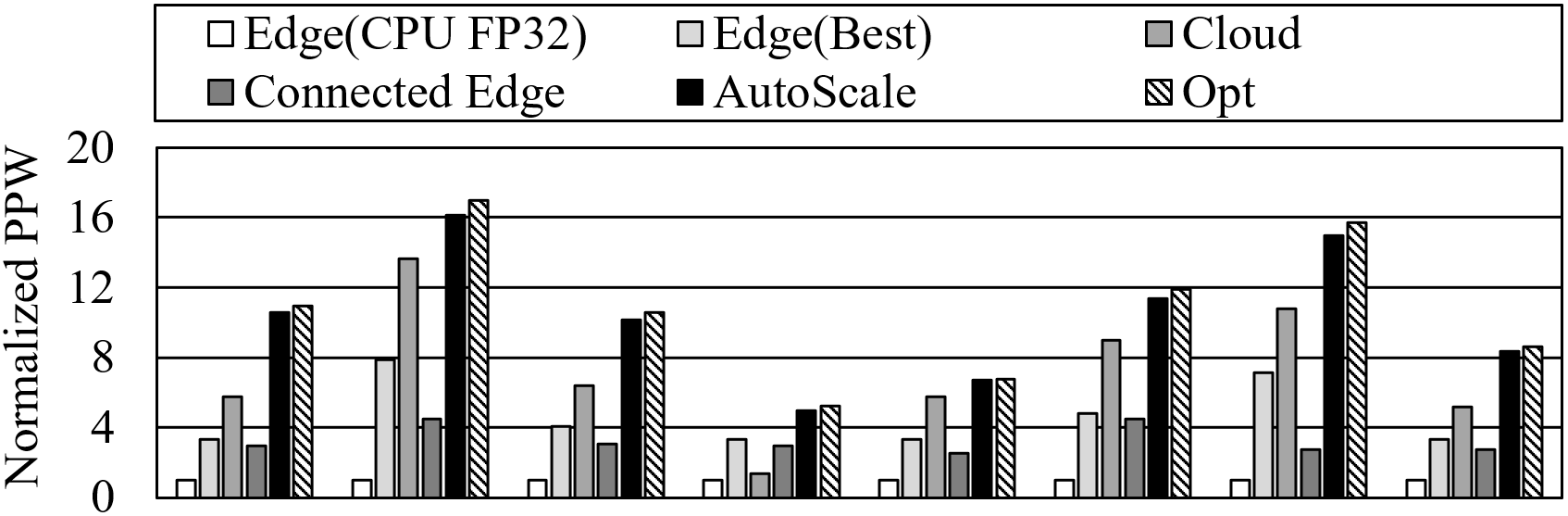}
    \includegraphics[width=\linewidth]{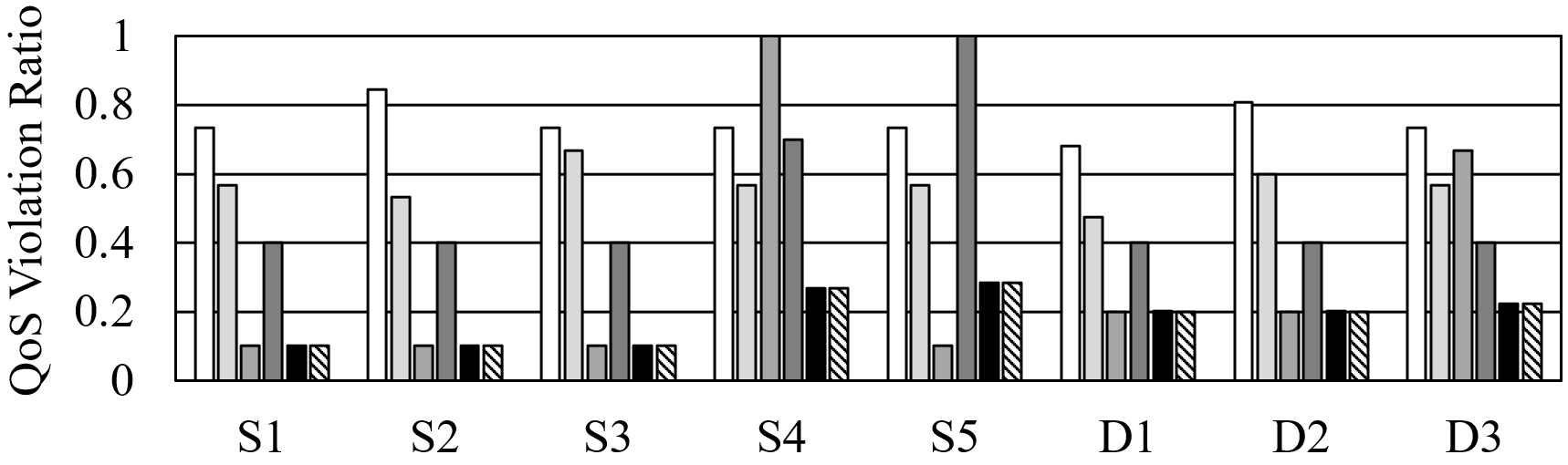}
    \vspace{-0.6cm}
    \caption{Since \textit{AutoScale} accurately predicts optimal target under the stochastic variances, it largely improves energy efficiency of DNN inference in realistic environments satisfying the QoS target as much as possible.}
    \label{fig:realistic environment}
    \vspace{-0.2cm}
\end{figure}

{\bf Adaptability to Stochastic Variances:} Figure \ref{fig:realistic environment} shows the average energy efficiency normalized to Edge(CPU FP32) and QoS violation ratio of DNN inference in the presence of stochastic variance. The x-axis represents the inference execution environments (Table \ref{table:Execution Environment}). Since \textit{AutoScale} accurately predicts the optimal execution scaling decision even in the presence of stochastic variance, it improves the average energy efficiency of DNN inference by 10.4X, 2.2X, 1.4X, and 3.2X, compared to Edge(CPU FP32), Edge(Best), Cloud, and Connected Edge, respectively, showing a similar QoS violation ratio as Opt.

\begin{figure}[t]
    \centering
    \includegraphics[width=\linewidth]{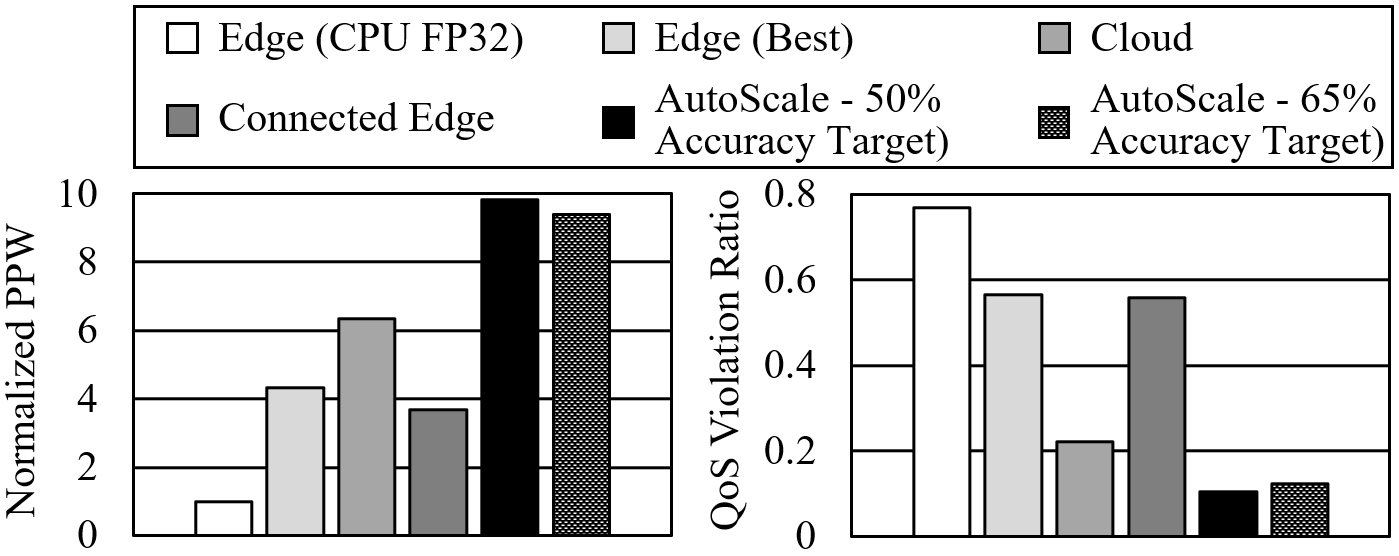}
    \vspace{-0.6cm}
    \caption{When \textit{AutoScale} uses higher accuracy target, its energy efficiency and QoS violation ratio are slightly degraded. Nevertheless, it still significantly improves energy efficiency compared to baselines.}
    \label{fig: accuracy target}
    \vspace{-0.3cm}
\end{figure}

{\bf Adaptability to Inference Quality Targets:} Figure \ref{fig: accuracy target} shows the average energy efficiency and the QoS violation ratio with different inference accuracy targets under \textit{AutoScale}. When \textit{AutoScale} uses 50\% as the inference accuracy target, it chooses processors on-device with low precision where some NN inference results in a low prediction accuracy. However, when \textit{AutoScale} uses higher inference accuracy target (i.e., 65\%), it does not choose the on-device processors with low precision operations. Due to this reason, when \textit{AutoScale} uses higher inference accuracy target, its energy efficiency and QoS violation ratio are slightly degraded. Nonetheless, it still improves the energy efficiency compared to the baseline.


\begin{figure}[t]
    \centering
    \includegraphics[width=0.93\linewidth]{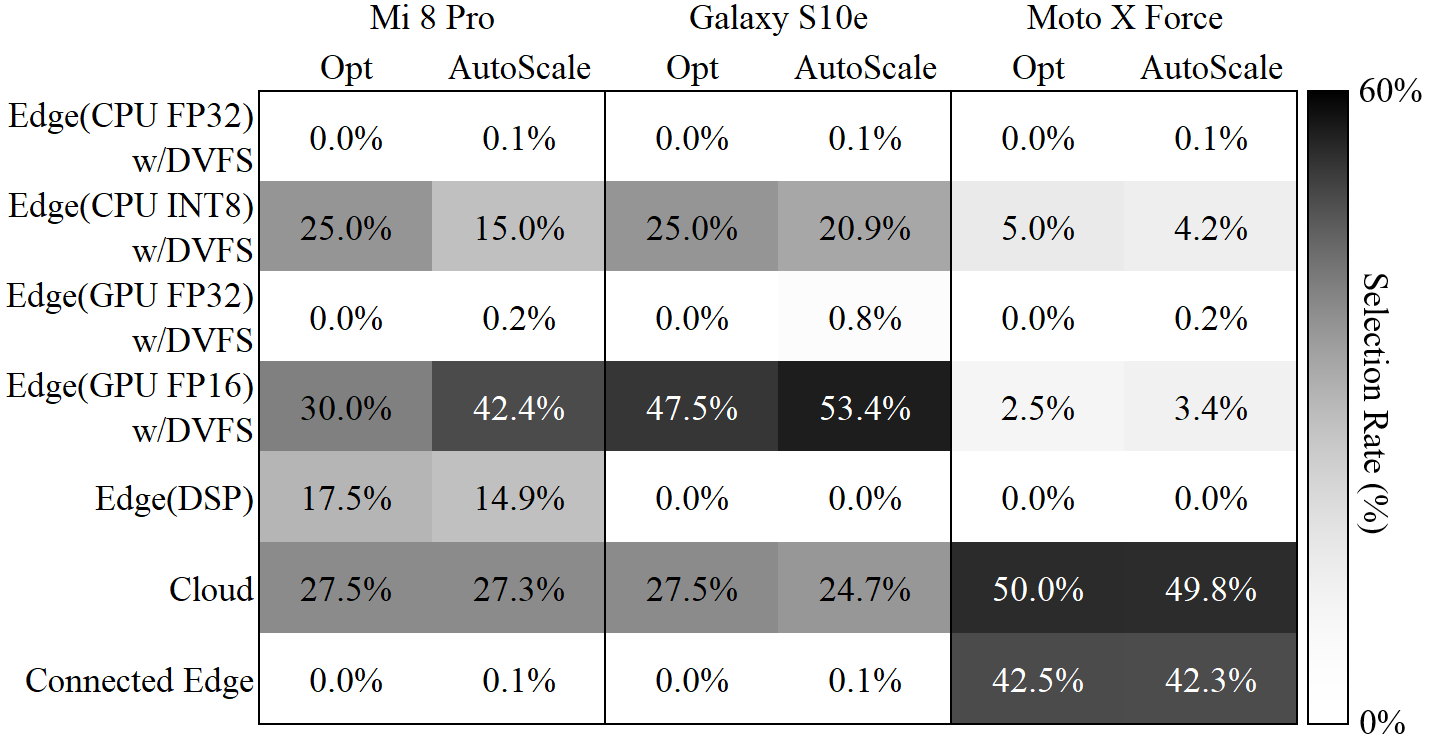}
    \vspace{-0.3cm}
    \caption{\textit{AutoScale} accurately selects the optimal execution target.}
    \label{fig:prediction}
    \vspace{-0.5cm}
\end{figure}

{\bf Prediction Accuracy:} To analyze the prediction accuracy of \textit{AutoScale}, we compare the execution scaling decision selected by \textit{AutoScale} to the optimal one. Figure \ref{fig:prediction} shows how the execution scaling decision is selected by \textit{AutoScale} and Opt on three mobile devices. 
\textit{AutoScale} accurately selects the optimal execution scaling decision for all devices, achieving 97.9\% of the prediction accuracy. \textit{AutoScale} mis-predicts the optimal execution target only when the energy difference between optimal execution target and the (mis-predicted) sub-optimal execution target is less than 1\%. This is because of the small error of $R_{energy}$. Although \textit{AutoScale} chooses the sub-optimal execution target for a few cases, it does not degrade the overall system energy efficiency and QoS violation ratio by much, as compared to Opt. This is due to the small energy difference between the optimal and sub-optimal ones.

\subsection{Overhead Analysis}
\label{sec:result5}

\begin{figure}[t]
    \centering
    \includegraphics[width=\linewidth]{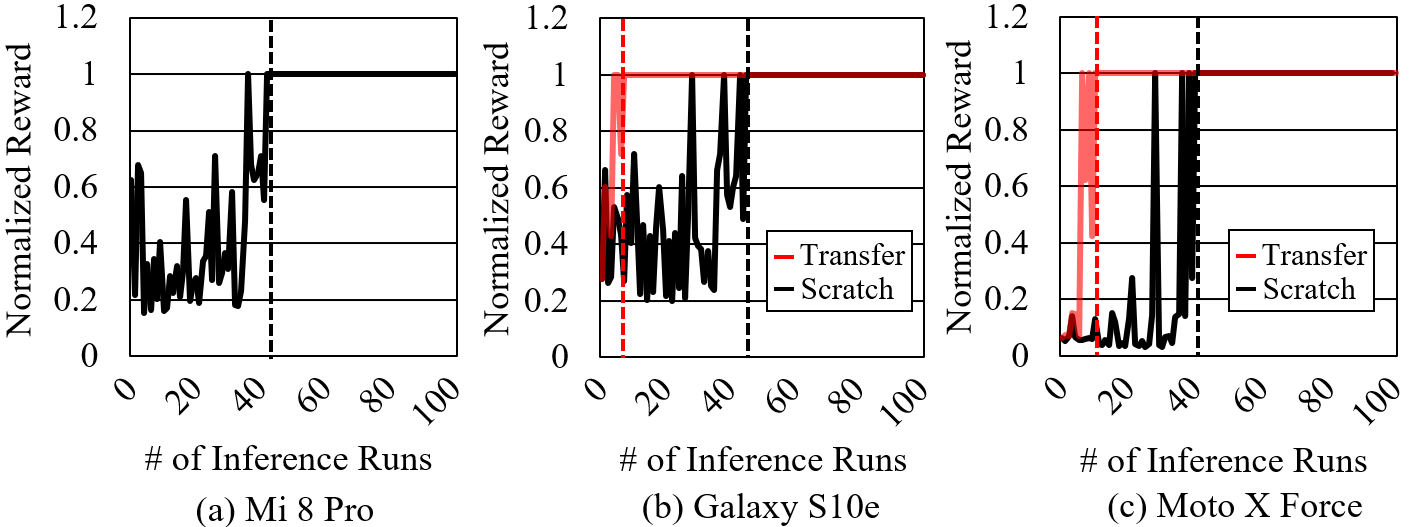}
    \vspace{-0.8cm}
    \caption{The reward is usually converged in 40-50 runs. Learning transfer can improve the speed of convergence.}
    \label{fig:train}
    \vspace{-0.5cm}
\end{figure}

{\bf Training Overhead:} Figure \ref{fig:train} shows that, when a model is trained from scratch, the reward converges with around 40-50 inference runs. Before the reward converges, compared to Opt., \textit{AutoScale} shows 18.9\% lower average energy efficiency. Nevertheless, it still achieves 66.1\% energy saving against Edge(CPU FP32). The training overhead can be alleviated with learning transfer. As shown in Figure~\ref{fig:train}, when the model trained on Mi8Pro is used for Galaxy S10e and Moto X Force, the training converges more rapidly, reducing the average training time overhead by 21.2\%. This result implies that \textit{AutoScale is able to capture and learn the common characteristics across the variety of edge inference workloads, performance and power profiles of edge systems, and uncertainties from the mobile-cloud environment}. 

{\bf Runtime Overhead:} To show viability for mobile inference deployment, we evaluate \textit{AutoScale} performance overhead. 
the average performance overhead of \textit{AutoScale} is 10.6 $\mu$s for Q-table training, which is 0.5\% of the lowest latency of mobile DNN inference. In addition, when using the trained Q-table, the overhead can be reduced to 7.3 $\mu$s, with only 0.3\% overhead. The energy overhead is only 0.4\% and 0.2\% of the total system energy consumption, respectively. The overall memory requirement of \textit{AutoScale} is 0.4MB, translating to only 0.01\% of the 3GB DRAM capacity of a typical mid-end mobile device~\cite{Motorola}.


\section {Related Work}
\label{sec:related work}

With the emergence of DNN-based intelligent services, energy optimization of mobile DNN inference has been widely studied. Due to the compute- and memory-intensive nature, many of the early works executed DNN inference in the cloud  \cite{YChen2019,ZFang2017,YGKim2017_2,YLiu2019}. As mobile systems become higher-performing~\cite{CGao15, MHalpern2016,LNHuynh2017,SWang2020_2}, there have been increasing pushes to execute DNN inference at the edge \cite{ECai2017,AEEshratifar2018,MHan2019,YKang2017,YKim2019,NDLane2016,SWang2020_1,SWang2020_2,CJWu2019,GZhong2019}. As an intermediate stage, many techniques tried to partition DNN inference execution between the cloud and local mobile device \cite{AEEshratifar2018,AEEshratifar2020,TGuo2018,YKang2017,GLiu2018}, based on performance/energy prediction models. However, these techniques do not consider fully executing inference at the edge. According to our analysis, there exist various cases where the edge inference execution outweighs the cloud inference execution by removing data transmission overhead. More importantly, the techniques also do not consider stochastic variances which largely affect the efficiency of inference execution.

To fully execute DNN inference at the edge, many optimizations, such as model architecture search \cite{MSandler2018,MTan2019,BZoph2018}, quantization \cite{MCourbariaux2016,JFromm2020,BJacob2018,JHKo2017,NDLane2016,CJWu2019,RZhao2019}, weight compression~\cite{CDing2017,SHan2016,DLi2018,SLiu2018} and graph pruning \cite{JYu2017,TZhang2018}, have been proposed. Along with these optimizations, deep learning compiler and programming stacks
have been improved to ease the adoption of energy efficient co-processors, such as GPUs, DSPs, and NPUs. On top of these works, many researchers tried to optimize the performance and/or energy efficiency of edge inference execution by exploiting the co-processors along with CPUs~\cite{ECai2017,MHan2019,YKim2019,NDLane2016,SWang2020_1,SWang2020_2,GZhong2019}. However, most of the above techniques are based on existing prediction approaches which are prone to being affected by stochastic variances. In addition, the above techniques also do not consider executing inference on connected systems, such as the cloud server or a locally connected mobile device.

Considering uncertainties in the mobile execution environment, various energy management techniques have been proposed~\cite{BGaudette2016,BGaudette2019,YGKim2017_2,YGKim2019,DShingari2018}. In order to maximize the energy efficiency of smartphones subject to user satisfaction demands under the memory interference, DORA takes a regression-based predictive approach to control the settings of mobile CPUs at runtime~\cite{DShingari2018}. Gaudette et al. proposed to use arbitrary polynomial chaos expansions to consider the impact of various sources of uncertainties on mobile user experience~\cite{BGaudette2019}. Other works explored the use of reinforcement learning to handle runtime variance for web browsers, for latency-critical cloud services, and for CPUs \cite{YChoi2019,SKMandal2020,RNishtala2017}. 

To the best of our knowledge, this is the first work that demonstrates the potential of machine learning inference at the edge by {\it automatically} leveraging programmable co-processors as well as other computing resources nearby and in the cloud. We examine a collection of machine learning-based predictive approach and tailor-design an automatic execution scaling engine with light-weight, customized reinforcement learning. {\it AutoScale} achieves near-optimal energy efficiency for DNN edge inference while taking into account \textit{stochastic variance}, particularly important for user quality of experience in the mobile domain.
\section{Conclusion}
\label{sec:conclusion}

Given the growing ubiquity of intelligent services, such as virtual assistance, face/image recognition, and language translation, deep learning inference is increasingly run at the edge. 
To enable energy-efficient inference at the edge, we propose an \textit{adaptive} and \textit{light-weight} deep learning execution scaling engine---\textit{AutoScale}. The in-depth characterization of DNN inference execution on mobile and edge-cloud systems demonstrates that the optimal scaling decision shifts depending on various features, namely NN characteristics, desired QoS and accuracy targets, underlying system profiles, and stochastic runtime variance. \textit{AutoScale} continuously learns and selects the optimal execution scaling decision by taking into account the features and dynamically adapting to the stochastic runtime variance. We design and construct representative edge inference use cases and mobile-cloud execution environment using off-the-shelf systems.
\textit{AutoScale} improves the energy efficiency of DNN inference by an average of 9.8X and 1.6X, as compared to the baseline settings of mobile CPU and cloud offloading, satisfying both the QoS and accuracy constraints. We demonstrate that \textit{AutoScale} is a viable solution and will pave the path forward by enabling future work on energy efficiency improvement for DNN edge inference in a variety of realistic execution environment. 

\section*{ACKNOWLEDGMENTS}
This work is supported in part by the National Science Foundation under \#1652132 and \#1525462.

\bibliographystyle{IEEEtranS}
\bibliography{refs}

\end{document}